\documentclass{article} 
\usepackage{iclr2027_conference,times}

\usepackage{amsmath,amsfonts,bm}

\def\eqref#1{equation~\ref{#1}}

\def\1{\bm{1}}

\DeclareMathAlphabet{\mathsfit}{\encodingdefault}{\sfdefault}{m}{sl}
\SetMathAlphabet{\mathsfit}{bold}{\encodingdefault}{\sfdefault}{bx}{n}

\usepackage{enumitem}

\usepackage[backref=page]{hyperref}
\usepackage{url}

\iclrfinalcopy

\usepackage{nicefrac}
 
\usepackage{ulem}
\usepackage{cleveref}
\usepackage{multirow}
\usepackage{tikz}

\usepackage{caption} 
\usepackage{microtype}

\usepackage{mathtools}

\usepackage{algorithm}
\usepackage{algpseudocode}
\usetikzlibrary{arrows.meta,positioning,backgrounds,fit}

\usepackage{booktabs}
\usepackage{graphicx}
\usepackage{array}

\usepackage{titlesec}
\titlespacing*{\paragraph}{0pt}{4pt}{1em}

\usepackage{amsmath,amssymb}
\usepackage{tabularx}
\usepackage{array}
\usepackage[T1]{fontenc}          
\usepackage[scaled=0.90]{helvet}  
\usepackage[table]{xcolor}
\newcommand{\hl}{\cellcolor[gray]{0.93}}
\usepackage[most]{tcolorbox}
\tcbuselibrary{theorems,skins}

\tcbuselibrary{listings,breakable}
\usepackage{listings}
\usepackage{bbm}

\definecolor{promptbg}{HTML}{F0F0F2}   

\definecolor{linkblue}{HTML}{1A5FB4}

\hypersetup{
  colorlinks=true,
  allcolors=linkblue
}

\lstdefinestyle{promptstyle}{
  basicstyle=\ttfamily\footnotesize\color{spark}\linespread{1.02}\selectfont,
  breaklines=true,
  breakatwhitespace=true,
  breakindent=0pt,
  breakautoindent=false,
  prebreak={}, postbreak={},
  columns=fullflexible,
  keepspaces=true,
  showstringspaces=false,
  upquote=true,
  aboveskip=0pt, belowskip=0pt,
}

\newtcblisting{promptplain}[1][]{
  listing only, listing style=promptstyle,
  enhanced, nobeforeafter,          
  colback=promptbg, colframe=promptbg, boxrule=0pt,
  arc=4pt, outer arc=4pt,
  left=8pt, right=8pt, top=6pt, bottom=6pt,
  before skip=10pt, after skip=10pt, #1}

\definecolor{bandgrey}{HTML}{F0F0F2}

\newtcolorbox{benchproblem}[1][]{
  enhanced, breakable, colback=defbg, colframe=defbg,
  boxrule=0pt, arc=4pt, outer arc=4pt,
  left=8pt, right=8pt, top=5pt, bottom=5pt,
  fonttitle=\sffamily\scriptsize\bfseries\scshape, coltitle=annot,
  attach title to upper=\par\vspace{2pt},
  fontupper=\small,
  before skip=10pt, after skip=10pt, #1}

\definecolor{wronganswer}{HTML}{B03A2E}
\definecolor{rightanswer}{HTML}{1E7A47}
\definecolor{bandgrey}{HTML}{F0F0F2}

\definecolor{defbg}{HTML}{F1F6FB}   
\definecolor{spark}{HTML}{2D2D2F}   
\definecolor{ink}{HTML}{2D2D2F}     
\definecolor{annot}{HTML}{5D5D5F}   
\definecolor{annot2}{HTML}{42658F}   

\newtcbtheorem[number within=section]{definitionbox}{Definition}{%
  enhanced, arc=4pt, outer arc=4pt,
  interior style={fill=defbg},
  frame empty, boxrule=0pt,
  colbacktitle=defbg, titlerule=0pt, toptitle=0pt, bottomtitle=0pt,
  fonttitle=\bfseries\scshape, coltitle=ink,
  separator sign none, description delimiters parenthesis, terminator sign={.},
  attach title to upper={\ },
  left=8pt, right=8pt, top=5pt, bottom=2pt,
  before skip=10pt, after skip=10pt,
}{def}

\newlength{\badgewidth}
\newlength{\numwidth}
\newcommand{\badge}[1]{%
  \makebox[\badgewidth][l]{%
    \makebox[\numwidth][r]{\normalfont\bfseries\color{spark}#1}\hspace{0.8em}}}

\newlength{\glossraise}
\newcommand{\gloss}[2][]{\raisebox{\glossraise}{\parbox[t]{\linewidth}{\raggedright\sffamily\footnotesize\linespread{1.18}\selectfont\leavevmode\color{annot2}%
  \ifx\\#1\\\else\mbox{\scshape\color{spark}#1}\enspace\fi\itshape #2}}}

\newenvironment{clauses}
  {\par\addvspace{5pt}\begingroup
   \setlength{\extrarowheight}{5pt}%
   \setlength{\abovedisplayskip}{4pt}\setlength{\belowdisplayskip}{4pt}%
   \setlength{\abovedisplayshortskip}{2pt}\setlength{\belowdisplayshortskip}{2pt}%
   \tabularx{\linewidth}{@{}>{\hangindent=\badgewidth\hangafter=1}X@{\hspace{1.1em}}
                       >{\raggedright\arraybackslash}p{0.32\linewidth}@{}}}
  {\endtabularx\endgroup\par\addvspace{2pt}}

\newtcolorbox{takeaway}[1][Summary]{%
  enhanced, arc=4pt, outer arc=4pt,
  interior style={fill=defbg},
  frame empty, boxrule=0pt,
  colbacktitle=defbg, titlerule=0pt, toptitle=0pt, bottomtitle=0pt,
  fonttitle=\bfseries, coltitle=ink,
  title={#1.}, attach title to upper={\ },
  left=8pt, right=8pt, top=3pt, bottom=3pt,
  before skip=10pt, after skip=10pt,
}

\newcolumntype{Y}{>{\centering\arraybackslash}X}

\crefname{figure}{Figure}{Figures}
\Crefname{figure}{Figure}{Figures}

\crefname{table}{Table}{Tables}
\Crefname{table}{Table}{Tables}

\crefname{algorithm}{Algorithm}{Algorithms}
\Crefname{algorithm}{Algorithm}{Algorithms}

\crefname{section}{Section}{Sections}
\Crefname{section}{Section}{Sections}

\newenvironment{points}
  {\par\addvspace{1.5pt}\begingroup
   \setlength{\parindent}{0pt}\setlength{\parskip}{2pt}}
  {\endgroup\par\addvspace{0pt}}
\newcommand{\point}[2]{\par\hangindent=\badgewidth\hangafter=1\badge{#1}#2\par}

\title{Towards Better Exploration in Sequential Test-Time Scaling}

\author{%
    \textbf{Joseph Rance$^{1,\dagger}$, Fabio Pizzati$^{2}$, Juil Sock$^{3}$, Woody Bayliss $^{3}$, Marc Górriz Blanch$^{3}$,} \\
    \textbf{Philip Torr$^{1}$, Adel Bibi$^{1}$} \\
    $^{1}$University of Oxford, $^{2}$MBZUAI, $^{3}$BBC R\&D \\
    $^{\dagger}$\texttt{joseph.rance@eng.ox.ac.uk}
}

\newcommand{\eg}{e.g.\ }
\newcommand{\ie}{i.e.\ }

\newcommand*\circled[1]{\tikz[baseline=(c.base)]{
  \node[shape=circle, draw, line width=0.4pt, inner sep=0.5pt, minimum size=1.1em] (c) {\small #1};}}

\begin{document}

\linepenalty=1000

\maketitle

\begin{abstract}
    Test-time scaling improves language model reasoning by spending additional compute at inference. However, both classes of existing methods often fail to continue improving over long timescales. \textit{Parallel} methods repeatedly sample independent answers from the model, scaling poorly on problems the model is unlikely to solve in a single attempt. In contrast, \textit{sequential} methods build on previous answers to access new ideas, yet so far have not been shown to reach answers beyond those found by parallel scaling. First, we show that sequential scaling often stops improving because it becomes prematurely trapped in an \textit{attractor}: a set of answers that prevents exploration of different answers once entered. Across 27 combinations of scaling methods, models, and benchmarks, we find that 53.8\% of sequential scaling trajectories enter an attractor within four iterations. Second, we show that a simple model-mixing intervention helps escape attractors. This reduces the attractor hit rate by 21.2 percentage points on average, expands solution coverage beyond a compute-matched \textit{parallel} baseline, and improves accuracy of recursive self-aggregation by at least 2.2 percentage points. Our results motivate refocusing long-horizon test-time scaling from parallel methods to sequential methods that improve previous answers.\footnote{Our code is available at: \url{https://anonymous.4open.science/r/tts-exploration}.}
\end{abstract}

\section{Introduction}

Test-time scaling has been shown to improve language model reasoning by using additional compute at test-time \citep{wu2024inference, snell2024scaling}. A widely used class of test-time scaling methods generates many independent answers to the given problem, and uses a heuristic to select the best \citep{zhang2025survey, wang2023selfconsistency}. However, these \textit{parallel scaling} methods scale poorly over long timespans: if they do not find a correct answer in the first few attempts, such an answer is unlikely to be accessible to the base model, leading to little improvement with increased scaling \citep{brown2024large}. Building methods to trade large amounts of compute for even a small improvement in reasoning capabilities would benefit high-value tasks such as developing large software systems \citep{jimenez2024swe} or scientific research \citep{science, novikov2025alphaevolve}.

A natural solution is to let the model iteratively refine its previous answers \citep{zhang2025survey, madaan2023self}. These \textit{sequential scaling} methods may be able to build on earlier answers to reach lines of reasoning that the model is unlikely to generate in an independent attempt. Yet there is so far little evidence of existing sequential scaling methods solving problems that parallel scaling could not \citep{huang2024large}. \textbf{In this paper, we explain why existing sequential scaling methods fail to improve beyond parallel scaling, and show that a simple intervention lets them reach answers that were not accessible through parallel scaling.}

\begin{figure}
    \vspace{-12pt}
    \centering
    \includegraphics[width=\linewidth]{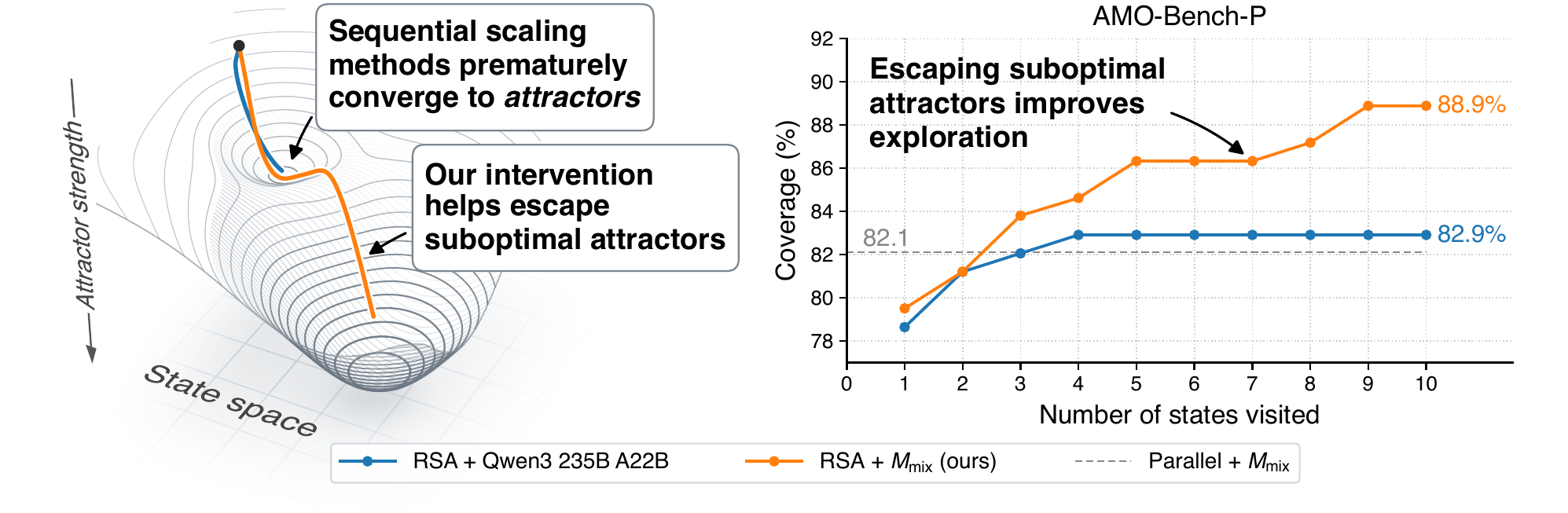}
    \vspace{-13pt}
    \caption{Escaping locally optimal attractors allows test-time scaling to improve for longer. \textbf{Left}: Schematic illustration of two sequential scaling trajectories through the state space, with minimum points as attractors. Both trajectories initially reach an attractor; the baseline (blue throughout this paper) remains trapped, while our intervention (orange throughout this paper) escapes and continues exploring. \textbf{Right}: Coverage (problems solved at any point during scaling) on AMO-Bench-P, over time. Our method continues improving after the baseline saturates by mixing Qwen3 235B A22B Thinking 2507 with GPT OSS 120B.}
    \label{intro:fig:main}
    \vspace{-6pt}
\end{figure}

Specifically, we propose that a common failure mode of sequential scaling is premature convergence to an \textit{attractor}: a set of answers that, once reached, the scaling method is unlikely to move away from. When stuck in an attractor, further scaling provides little benefit because it is prevented from exploring new answers. This behaviour can be useful for sequential scaling because we want states that contain \textit{correct} answers to be attractors. However, we hypothesise, as the blue line in \cref{intro:fig:main} (left) illustrates, that sequential scaling methods prematurely converge to attractors \textit{before} finding a correct answer, preventing further iterations from improving towards a correct answer. We test this hypothesis across 27 combinations of diverse sequential scaling methods, models, and reasoning-focused benchmarks, finding that an average of 53.8\% of trajectories get stuck in an attractor within the first four iterations (mean of \cref{attractors:tab:part_1}).

As a simple intervention to demonstrate the potential of sequential scaling, we propose repeatedly switching the underlying language model to escape these attractors. Our intuition is that an attractor for one model may not be an attractor for another, allowing the second model to explore new ideas when the first hits an attractor. For example, the orange line in \cref{intro:fig:main} (left) initially enters the same suboptimal attractor as the blue trajectory, but switching models allows it to escape and continue towards the correct answer. \Cref{intro:fig:main} (right) shows that this improves sequential scaling to reach answers not found by parallel scaling.

In short, the primary takeaway of this paper is to demonstrate the potential of sequential scaling, which motivates refocusing long-horizon test-time scaling from parallel methods (a search process under a fixed model policy) to sequential methods that improve on previous answers (a learning process where the model iteratively builds on its ideas to improve its search). Our contributions are the following:
\vspace{-1pt}
\begin{enumerate}[topsep=0pt,itemsep=0pt]

    \item We introduce a general formulation for test-time scaling methods, and the notion of attractors within this formulation.

    \item We show that existing sequential scaling methods prematurely converge to attractors: across 27 scaling-method--model--benchmark combinations, 53.8\% of trajectories enter an attractor within the first four states, limiting further gains in the number of problems solved.

    \item We show that a simple model-mixing intervention lets existing sequential scaling methods escape these attractors, access 3.3 percentage points more correct answers than parallel scaling, and improve benchmark performance by 2.2 percentage points.

\end{enumerate}

\section{Related Work}\label{sec:rel}

\subsection{Parallel scaling}

Test-time scaling is often categorised into parallel methods and sequential methods \citep{zhang2025survey}. Parallel scaling methods have been shown to be a compute-efficient option for improving language model output quality \citep{wu2024inference}. These methods sample $N$ independent answers from the model, and then try to select the best. The idea is to improve the probability that the model produces a good answer by considering more than one high-probability answer. For example, \cite{wang2023selfconsistency} return the most common final answer out of $N$. Other parallel scaling methods use more advanced approaches to select between the $N$ answers \citep{lightman2024lets, brown2024large}, or generate higher quality and more diverse answers \citep{jiang2023llm, yao2023tree}.

However, for problems that the base model is not already likely to solve,\footnote{which may require more innovative solutions, and therefore also be the most interesting, \eg for recursive self-improvement \citep{chen2026recursive, zelikman2022star}.} parallel methods must scale up an expensive search procedure to obtain at least one correct answer out of $N$. Consequently, once the `easy' problems are exhausted, each new problem solved often requires exponentially more test-time compute \citep{brown2024large, wang2025scaling, chen2026does}, leading to a performance plateau under practical budgets. For test-time scaling methods to continue to improve, they must devote additional compute to individual answers.

\subsection{Sequential scaling}\label{sec:rel:sequential}

Sequential scaling allocates additional test-time compute to a single reasoning chain by generating new answers conditioned on previous answers. Each new answer aims to build on the `understanding' encoded in the previous, potentially shifting the answer distribution away from that of parallel scaling. For example, self-refine \citep{madaan2023self} starts with one answer, and then generates each new one using the previous answer and a language-model-generated critique of that answer as context. There are many approaches to stimulate the generation of new answers \citep{shinn2023reflexion, yan2026inftythink, snell2024scaling}, which are often combined with parallel scaling to form \textit{hybrid} scaling \citep{liang2024encouraging, zhou2026multi}. An example of the latter is recursive self-aggregation \citep{venkatraman2025recursive}, which iterates on a state of $N$ answers, where each answer is conditioned on $K$ random answers from the previous state, and the output is selected from the last state.

However, none of these works demonstrate that sequential scaling with off-the-shelf models generates answers that are better than those generated by parallel scaling. In fact, existing evidence points in the opposite direction: \cite{huang2024large} show that sequential scaling often fails to produce \textit{any} benefit \citep[see also][]{kamoi2024can, stechly2025self, wang2025scaling, zeng2025revisiting, choi2025debate, mang2026humansstillbeatagents}.

Previous work such as that of \citet{liang2024encouraging} has observed that specific instances of test-time scaling may fail due to repeating the same answers rather than exploring. There are other works that similarly observe attractors in specific cases \citep[][]{ko2026attractor, chen2026diversity, gurkan2026mutation, wang2025unveiling, tacheny2025geometric, awadhiya2026reasoning, wu2026language, huang2026, fein-ashley2026solve}. In contrast, we claim that attractors are a common failure mode across the general class of sequential scaling methods.

\subsection{Mixtures of models}\label{sec:rel:mix}

Prior work has studied whether mixing different language models within sequential scaling improves performance, with mixed results. Several works report that mixing models improves performance for specific methods: \citet{zhang2025stop}, \citet{chen2024reconcile}, and \citet{yang2026understanding} demonstrate gains in multi-agent debate, while \citet{wang2024} and \citet{yang2026multi} report improvements for other hybrid approaches. Meanwhile, \citet{wang2025generalizing} find that mixing models often \textit{underperforms} the strongest constituent model, with similar negative results reported by \citet{li2024rethinking}. While these works compare overall performance relative to specific sequential scaling methods, we focus on whether our model-mixing intervention, when applied to a \textit{range} of existing sequential methods, escapes attractors and increases \textit{exploration} beyond that achieved through parallel scaling.

\section{A General Formulation of Test-time Scaling}

In this section, we unify test-time scaling methods under a general formulation that allows us to reason about their shared properties through the lens of a dynamical system. We focus on test-time scaling methods that operate entirely through natural language using off-the-shelf models without task-specific feedback (further discussed in Appendix~\ref{sec:app:scope}).

Given a question $X$ and language model $M$, we aim to generate an answer $Y$. For any prompt text $p$, we can sample responses from the model's output distribution $M(p)$. We define a test-time scaling method to consist of three functions $(f_s, f_r, f_o)$, which maintain a state $S$, and return an answer $Y_T$ after $T$ states.

\noindent\textbf{(1) Initial state generator.} $f_s$ generates the first state $S_1$ from $X$. We define a state $S_t$ to be a length-$N$ vector of complete responses generated from the language model: $S_t=(S_t^1, S_t^2, \ldots, S_t^N)$, where $S_t^{i}\sim M(\ldots)$. For example, a sequential method that iterates on only the most recent answer would have $N=1$, with $S_t^1$ being the answer at timestep $t$.
    \begin{equation}
        S_1\sim f_s(X,M)
    \end{equation}
\noindent\textbf{(2) Next-state transition.} $f_r$ generates $S_{t+1}$ from $S_t$. We iteratively apply $f_r$ for $T-1$ iterations, until we have $S_T$. For example, in self-refine \citep{madaan2023self}, each application of $f_r$ first generates a critique of the current state $C\sim M($`Critique $S_t$ for $X$'$)$, and then conditions the next state on this critique, $S_{t+1}^1\sim M($`Revise $S_t$ for $X$ using $C$'$)$.
    \begin{equation}
        S_{t+1}\sim f_r(S_t;X,M)
    \end{equation}
\noindent\textbf{(3) Output function.} A state $S_t$ cannot be returned on its own, because it does not necessarily contain a single answer. Instead, we define $f_o$ to map from states to answers. For example, in recursive self-aggregation, we first sample $i\sim\mathcal{U}([N])$, and then return $Y_t=S_t^{i}$.
    \begin{equation}
        Y_t\sim f_o(S_t;X,M)
    \end{equation}

A trajectory is a sequence of sampled states $(S_1, S_2, \ldots, S_T)$. The goal is to build a test-time scaling method where each answer $Y_t$ is more likely to be correct than the previous answer $Y_{t-1}$, so that we can use extra compute to get a better answer. \Cref{back:tab:methods} shows how different test-time scaling methods discussed in \cref{sec:rel} fit into the proposed formulation. For example, on the third row, reasoning cache \citep{wu2026reasoning} uses $f_s$ to initialise an answer $S_1^1$ and an empty summary $S_1^2$, $f_r$ to generate a new summary based on the previous state and a new answer conditioned on this summary, and $f_o$ to return the current answer.

\begin{table}[t]
\centering
\caption{How different test-time scaling methods fit into our general formulation.}
\label{back:tab:methods}
\vspace{2pt}
\newsavebox{\ttsdiagrambox}
\newsavebox{\ttsrowbox}
\resizebox{\textwidth}{!}{%
\vbox{%
\offinterlineskip
\tikzset{
  stn/.style  = {circle, draw=blue!55!black,   fill=blue!12,   minimum size=0.46cm, inner sep=0pt, font=\tiny},
  smn/.style  = {circle, draw=orange!75!black, fill=orange!22, minimum size=0.46cm, inner sep=0pt, font=\tiny},
  res/.style  = {rounded corners=1pt, draw=green!45!black, fill=green!14,
                 minimum width=0.52cm, minimum height=0.30cm, inner sep=1pt, font=\tiny},
  ar/.style   = {-{Stealth[length=3pt,width=2.4pt]}, black!60, line width=0.35pt},
  arf/.style  = {-{Stealth[length=2.8pt,width=2.3pt]}, black!50, line width=0.3pt},
  cell/.style = {anchor=west, align=left, font=\scriptsize, inner sep=1pt},
  hdr/.style  = {cell, font=\scriptsize\bfseries},
  meth/.style = {cell, font=\scriptsize\bfseries},
  lbl/.style  = {font=\tiny, text=black!60, inner sep=0.5pt},
  ell/.style  = {font=\tiny, text=black!55, inner sep=1pt, fill=white},
}
\hyphenpenalty=10000\relax
\def\W{14.95}
\def\cA{0.05}\def\cB{2.15}\def\cD{5.75}\def\cE{7.60}\def\cF{12.30}
\newcommand{\subttl}[1]{\\[-1.5pt]{\tiny\mdseries\itshape\color{black!55}#1}}
\def\xA{2.48}\def\xB{3.23}\def\xC{3.98}\def\xE{4.53}\def\xY{5.16}

\def\ttsrowpad{0.10cm}
\newcommand{\ttsplacerow}[3]{%
  \sbox{\ttsrowbox}{%
    \begin{tikzpicture}
      \path (0,0) (\W,0);
      #3
    \end{tikzpicture}%
  }%
  \vskip\ttsrowpad
  \hbox{\usebox{\ttsrowbox}}%
  \vskip\ttsrowpad
  {\color{#2}\hrule height #1 width \W cm}%
}
\newcommand{\ttsrow}[5][center]{%
  \sbox{\ttsdiagrambox}{%
    \begin{tikzpicture}
      \path (0,0) (\W,0);
      #4
    \end{tikzpicture}%
  }%
  \ttsplacerow{#2}{#3}{%
    #5
    \def\ttsdiagramy{0pt}%
    \def\ttsdiagramalign{#1}%
    \def\ttstopalign{top}%
    \ifx\ttsdiagramalign\ttstopalign
      \pgfextracty{\dimen0}{\pgfpointanchor{current bounding box}{north}}%
      \pgfmathsetlengthmacro{\ttsdiagramy}{%
        max(0pt,\the\dimen0-(\ht\ttsdiagrambox+\dp\ttsdiagrambox)/2)}%
    \fi
    \node[anchor=west,inner sep=0pt,outer sep=0pt]
      at (0,\ttsdiagramy) {\usebox{\ttsdiagrambox}};
  }%
}

\hrule height 0.9pt width \W cm
\ttsplacerow{0.5pt}{black}{%
\node[hdr] at (\cA,0) {Method};
\node[hdr] at (\cB,0) {Structure};
\node[hdr] at (\cD,0) {Start $f_s$};
\node[hdr] at (\cE,0) {Transition $f_r$};
\node[hdr] at (\cF,0) {Output $f_o$};
}

\ttsrow{0.2pt}{black!30}{%
\node[stn] (a1) at (\xA,0) {$S_1$};
\node[res] (ay) at (\xY,0) {$Y_T$};
\draw[ar] (a1) -- (ay);
}{%
\node[meth,text width=1.95cm] at (\cA,0) {Direct prompt\subttl{no scaling}};
\node[cell,text width=1.60cm] at (\cD,0) {$S_1^1\sim M(X)$};
\node[cell,text width=4.65cm] at (\cE,0) {---};
\node[cell,text width=2.60cm] at (\cF,0) {$Y_t=S_t^{1}$};
}

\ttsrow{0.2pt}{black!30}{%
\node[stn] (b1) at (\xA,{0+0.60}) {$S_1^1$};
\node[stn] (b2) at (\xA,0)        {$S_1^2$};
\node[stn] (b3) at (\xA,{0-0.60}) {$S_1^N$};
\node[res] (by) at (\xY,0) {$Y_T$};
\draw[ar] (b1) -- (by);
\draw[ar] (b2) -- (by);
\draw[ar] (b3) -- (by);
\node[lbl,fill=white] at (3.65,0) {majority vote};
}{%
\node[meth,text width=1.95cm] at (\cA,0) {Self-consistency \citep{wang2023selfconsistency}\subttl{parallel}};
\node[cell,text width=1.60cm] at (\cD,0) {$S_1^{i}\sim M(X)$};
\node[cell,text width=4.65cm] at (\cE,0) {---};
\node[cell,text width=2.60cm] at (\cF,0) {\mbox{$Y_t=\mathrm{MajVote}(S_t)$}};
}

\ttsrow{0.2pt}{black!30}{%
\node[stn] (d1) at (\xA,{0+0.36}) {$S_1^1$};
\node[stn] (d2) at (\xB,{0+0.36}) {$S_2^1$};
\node[stn] (d3) at (\xC,{0+0.36}) {$S_3^1$};
\node[smn] (e2) at (\xB,{0-0.36}) {$S_2^2$};
\node[smn] (e3) at (\xC,{0-0.36}) {$S_3^2$};
\node[res] (dy) at (\xY,{0+0.36}) {$Y_T$};
\draw[ar] (d1) -- (e2); \draw[ar] (e2) -- (d2);
\draw[ar] (d2) -- (e3); \draw[ar] (e2) -- (e3); \draw[ar] (e3) -- (d3);
\draw[ar] (d3) -- (dy);
\node[ell] at (\xE,{0+0.36}) {$\cdots$};
\node[lbl] at (3.605,{0-0.78}) {summarise};
\node[lbl] at (4.43,0) {answer};
}{%
\node[meth,text width=1.95cm] at (\cA,0) {Reasoning Cache \citep{wu2026reasoning}\subttl{sequential}};
\node[cell,text width=1.60cm] at (\cD,0) {$S_1^{1}\sim M(X)$\\[1pt] $S_1^{2}=\varnothing$};
\node[cell,text width=4.65cm] at (\cE,0) {%
  $S_{t+1}^{2}\sim M($`Summarise $S_t$ for $X$'$)$\\[1pt]
  $S_{t+1}^{1}\sim M($`Answer $X$ given $S_{t+1}^{2}$'$)$};
\node[cell,text width=2.60cm] at (\cF,0) {$Y_t=S_t^{1}$};
}

\ttsrow[top]{0.2pt}{black!30}{%
\node[stn] (c1) at (\xA,0) {$S_1$};
\node[stn] (c2) at (\xB,0) {$S_2$};
\node[stn] (c3) at (\xC,0) {$S_3$};
\node[res] (cy) at (\xY,0) {$Y_T$};
\draw[ar] (c1) -- (c2);
\draw[ar] (c2) -- (c3);
\draw[ar] (c3) -- (cy);
\node[ell] at (\xE,0) {$\cdots$};
\node[lbl,align=center] at (2.855,{0+0.44}) {critique\\then revise};
}{%
\node[meth,text width=1.95cm] at (\cA,0) {Self-Refine \citep{madaan2023self}\subttl{sequential}};
\node[cell,text width=1.60cm] at (\cD,0) {$S_1^1\sim M(X)$};
\node[cell,text width=4.65cm] at (\cE,0) {%
  $C\sim M($`Critique $S_t$ for $X$'$)$\\[1pt]
  $S_{t+1}^1\sim M($`Revise $S_t$ for $X$ using $C$'$)$};
\node[cell,text width=2.60cm] at (\cF,0) {$Y_t=S_t^{1}$};
}

\ttsrow{0.9pt}{black}{%
\node[stn] (f1) at (\xA,{0+0.60}) {$S_1^1$};
\node[stn] (f2) at (\xA,0)        {$S_1^2$};
\node[stn] (f3) at (\xA,{0-0.60}) {$S_1^N$};
\node[stn] (g1) at (\xB,{0+0.60}) {$S_2^1$};
\node[stn] (g2) at (\xB,0)        {$S_2^2$};
\node[stn] (g3) at (\xB,{0-0.60}) {$S_2^N$};
\node[stn] (h1) at (4.25,{0+0.60}) {$S_T^1$};
\node[stn] (h2) at (4.25,0)        {$S_T^2$};
\node[stn] (h3) at (4.25,{0-0.60}) {$S_T^N$};
\node[res] (gy) at (\xY,0) {$Y_T$};
\draw[arf] (f1) -- (g1); \draw[arf] (f1) -- (g2);
\draw[arf] (f2) -- (g1); \draw[arf] (f2) -- (g3);
\draw[arf] (f3) -- (g2); \draw[arf] (f3) -- (g3);
\draw[arf] (g1) -- (h1); \draw[arf] (g1) -- (h2);
\draw[arf] (g2) -- (h1); \draw[arf] (g2) -- (h3);
\draw[arf] (g3) -- (h2); \draw[arf] (g3) -- (h3);
\fill[white,overlay] (3.610,{0-0.95}) rectangle (3.870,{0+0.95});
\node[ell] at (3.740,0) {$\cdots$};
\foreach \n in {h1,h2,h3}{\draw[ar] (\n) -- (gy);}
\node[lbl] at (3.44,{0+1.00}) {aggregate $K$};
\node[lbl] at (5.10,{0+0.40}) {rand.\ $j$};
}{%
\node[meth,text width=1.95cm] at (\cA,0) {Recursive Self-Aggregation \citep{venkatraman2025recursive}\subttl{hybrid}};
\node[cell,text width=1.60cm] at (\cD,0) {$S_1^{i}\sim M(X)$};
\node[cell,text width=4.65cm] at (\cE,0) {%
  $J\sim\mathcal{U}\big(\{I\subseteq[N]:|I|=K\}\big)$\\[1pt]
  $A_t^{i}=\{S_t^{j}\}_{j\in J}$\\[1pt]
  $S_{t+1}^{i}\sim M($`Answer $X$ given $A_t^{i}$'$)$};
\node[cell,text width=2.60cm] at (\cF,0) {\mbox{$Y_t=S_t^{i}$,} \mbox{$i\sim\mathcal{U}([N])$}};
}

}%
}
\vspace{-11pt}
\end{table}

\section{Why Sequential Scaling Fails}\label{sec:attractors}

We now study why sequential scaling methods fail to outperform parallel methods, using our formulation of sequential scaling methods as a dynamical system. In \cref{sec:attractors:hyp}, we present our hypothesis: sequential scaling is often prevented from making further progress because it gets stuck in a suboptimal attractor. We test this hypothesis in \cref{sec:attractors:evidence}.

\subsection{Hypothesis: sequential scaling gets stuck in suboptimal attractors}\label{sec:attractors:hyp}

Our hypothesis stems from the observation that some sets of states generated during sequential scaling represent attractors: sets that, once reached, the scaling method is unlikely to move away from (\eg a set $A$ where if $S_t\in A$ then $S_{t+1}\in A$). We hypothesise that a common failure mode in sequential scaling comes from getting stuck, within the first few states of a trajectory, at an attractor that does not contain a correct answer, preventing further exploration. Such attractors could arise if the language model $M$ is biased towards reusing the reasoning in previous answers included in its prompt \citep{nehring2024large, ali2026mitigating, holtzmancurious}, stopping us from escaping a local optimum. In the next section (\ref{sec:attractors:evidence}), we test this hypothesis, identifying attractors using Definition~\ref{def}.

\begin{definitionbox}{Attractor}{attractor}\label{def}
For $0<\epsilon_0,\epsilon_1<1$, a non-empty set of states $A$ is an attractor iff
\vspace{-3pt}
\begin{clauses}
  \badge{i}for any $s\in A$,
  \mbox{$\sum_{s^\prime\not\in A}P(S_{i+1}=s^\prime|S_i=s)<\epsilon_0$};
  & \gloss{The next state is unlikely to fall outside the attractor.} \\[11pt]
  \badge{ii}there is a set $B(A)$ where $A\subset B(A)$, and, for all
  $s\in B(A)$ and some $k>0$, \mbox{$P(S_{i+k}\in A|S_i=s)>1-\epsilon_1$};
  & \gloss{States in the basin of attraction transition into the attractor.} \\[5pt]
  \badge{iii}no $A^\prime\subset A$ is an attractor.
  & \gloss{Only accept minimal attractors.}
\end{clauses}
\end{definitionbox}

\subsection{Evidence that sequential scaling gets stuck in suboptimal attractors}\label{sec:attractors:evidence}

We now provide empirical evidence for the hypothesis in \cref{sec:attractors:hyp}, across 27 combinations of sequential scaling methods, language models, and reasoning-focused benchmarks.

\paragraph{Experimental setup.} We focus our experiments on benchmarks that involve maths and games, where success depends on reasoning rather than extensive domain knowledge: \textbf{(1) IMO-AnswerBench (IMO):} 400 questions from mathematical olympiad competitions \citep{luong-etal-2025-towards}; \textbf{(2) AMO-Bench-P (AMO):} 39 maths questions that are less saturated than those of IMO-AnswerBench \citep{an2025amobench}; \textbf{(3) Reasoning Gym Games (RGG):} 100 questions that require reasoning about games such as chess, generated by \cite{venkatraman2025recursive} from the Reasoning Gym benchmark \citep{stojanovski2026reasoning}. In all three benchmarks, the test-time scaling method receives a question $X$ as text. Its output $Y_T$ should include a \textit{final answer}, such as a single number or mathematical expression. This final answer is checked for equivalence against the golden answer.

We test the three sequential scaling methods described in \cref{back:tab:methods}: \textbf{(1)} \textbf{Reasoning cache} \citep{wu2026reasoning}, as a simple configuration for chain-of-thought with compaction (we do not fine-tune, unlike \citet{wu2026reasoning}); \textbf{(2)} \textbf{Self-refine} \citep{madaan2023self}, as a critique-based approach; \textbf{(3)} \textbf{Recursive self-aggregation (RSA)} \citep{venkatraman2025recursive} as a hybrid approach with strong empirical results. Compaction and critique cover the most popular works on pure sequential scaling \citep{zhang2025survey}, while RSA tests whether the same dynamics persist when combined with parallel scaling.

In our experiments, we generate $T=60$ states per trajectory for reasoning cache and self-refine, and $T=5$ for RSA. We use $N=12$ answers per state in RSA, and $N=60$ answers for our parallel baselines, so that all methods are matched in their total number of attempts to solve the problem. In \cref{intro:fig:main,mix:fig:number,mix:fig:final}, these $T$ values are doubled, to demonstrate capabilities over long timescales. We test each method with each of the models: GPT OSS 120B (medium reasoning effort) \citep{openai2025gptoss}, Gemma 3 12B IT \citep{gemmateam2025gemma3}, Granite 4.1 8B \citep{granite2026}. We select these models to cover a diverse set of model families and capabilities, while providing a fair comparison with our results in \cref{sec:mix}.

We identify attractors by grouping semantically similar states using their text embeddings, and then applying Definition~\ref{def} to the Markov chain over the resulting groups \citep{prinz2011markov}. Appendix~\ref{sec:app:exp} further details our scaling methods, models, and benchmarks, how we assign grades and attractors, and our validation of the resulting attractors.

\begin{table}[t]
\centering
\caption{\textbf{Left columns}: Probability that each scaling method hits an attractor within the first 4 states. Higher values mean the scaling-method--model combination is more likely to get stuck early on. \textbf{Right columns}: Probability that, if two trajectories -- using the same scaling method, model, and question -- both terminate in attractors, then exactly one of the attractors contains only incorrect answers, making it a suboptimal attractor.}
\label{attractors:tab:part_1}
\vspace{2pt}
\begin{tabularx}{\textwidth}{llYYY YYY}
\toprule
& &
\multicolumn{3}{c}{Attractor hit by $S_4$ (\%) $\downarrow$} &
\multicolumn{3}{c}{Different correctness (\%) $\downarrow$} \\
\cmidrule(lr){3-5}
\cmidrule(lr){6-8}
Method & Model
& IMO & AMO & RGG
& IMO & AMO & RGG \\
\midrule

\multirow{3}{*}{\parbox{2.8cm}{Reasoning Cache}}
& GPT OSS 120B
& 50.1 & 39.3 & 23.3
& 15.1 & 21.0 & 9.5 \\
& Gemma 3 12B IT
& 48.0 & 37.6 & 23.7
& 17.1 & 19.4 & 20.4 \\
& Granite 4.1 8B
& 73.2 & 62.4 & 35.3
& 22.2 & 20.1 & 27.4 \\

\midrule

\multirow{3}{*}{\parbox{2.8cm}{Self-Refine}}
& GPT OSS 120B
& 57.1 & 37.6 & 27.0
& 15.5 & 14.3 & 10.5 \\
& Gemma 3 12B IT
& 48.0 & 33.3 & 33.3
& 16.7 & 17.4 & 10.3 \\
& Granite 4.1 8B
& 71.4 & 60.7 & 37.7
& 28.1 & 27.9 & 15.5 \\

\midrule

\multirow{3}{*}{\parbox{2.8cm}{Recursive\\Self-Aggregation}}
& GPT OSS 120B
& 88.7 & 72.6 & 81.0
& 12.7 & 9.8 & 12.7 \\
& Gemma 3 12B IT
& 60.7 & 61.5 & 77.3
& 5.1 & 7.3 & 7.3 \\
& Granite 4.1 8B
& 69.3 & 60.7 & 81.3
& 10.9 & 10.5 & 11.3 \\

\bottomrule
\end{tabularx}

\end{table}

\paragraph{Finding 1: Sequential scaling gets stuck in attractors after visiting a small number of states.} \Cref{attractors:tab:part_1} (left) shows the proportion of trajectories where at least one of the first four states is identified as in an attractor (\ie hit an attractor by $S_4$). 23.3–88.7\% of trajectories enter an attractor within the first four states, halting exploration to other states. This attractor effect is clearest for RSA, which we attribute to including more answers in each state, increasing the number of opportunities to find an attractor quickly.

\paragraph{Finding 2: The attractors sequential scaling gets stuck in can be suboptimal.} Getting stuck in an attractor is only harmful if the attractor contains only incorrect answers, and a correct answer could have otherwise been found. To measure how often this occurs, \cref{attractors:tab:part_1} (right) shows the probability that, if two trajectories -- using the same scaling method, model, and question -- both terminate in attractors, then exactly one of the attractors contains only incorrect answers. On average, 15.4\% of these trajectory pairs differ in correctness, implying that \textit{at least} $\left. 15.4 \middle/ 2\right.=7.7\%$ of attractors must be \textit{suboptimal} (\ie incorrect even though we have evidence -- from the other attractor in the pair -- that the model could reach a correct answer). Given that 74.7\% of trajectories end in attractors, if we could escape suboptimal attractors, we could improve accuracy by \textit{at least} $74.7\%\times7.7\approx5.8$ percentage points. In Appendix~\ref{sec:app:corr}, we show the presence of even more (potentially-suboptimal) attractors.

\begin{takeaway}
\vspace{-2pt}
\begin{points}
  \point{i}{The sequential scaling methods tested converge to suboptimal attractors;}
  \point{ii}{They get stuck at these attractors rather than exploring;}
  \point{iii}{This constrains answer correctness to that of the suboptimal attractor.}
\end{points}
\end{takeaway}

\section{Improved Exploration by Escaping Suboptimal Attractors}\label{sec:mix}

We now present a simple model-mixing intervention to help sequential scaling methods escape attractors, with the aim of letting these methods solve problems that are not solved by parallel methods.

\subsection{Hypothesis: sequential scaling can escape attractors by mixing models}

\begin{figure}
    \centering
    \includegraphics[width=0.9\linewidth]{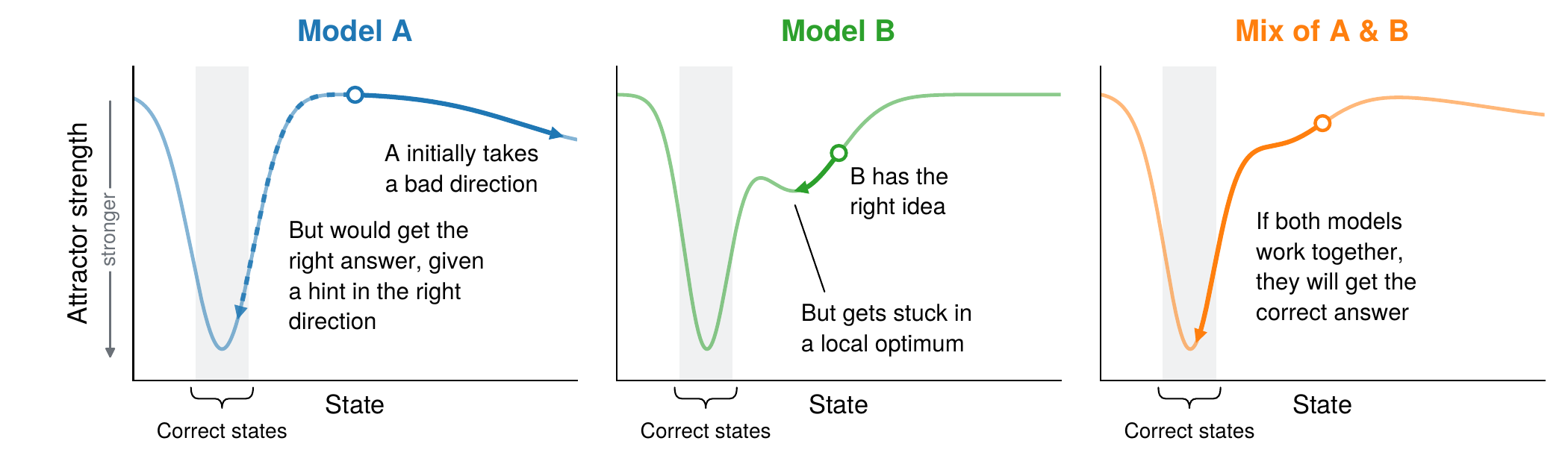}
    \caption{Illustration of how mixing models can help escape suboptimal attractors. Y axis represents attractor strength (\ie attractors that satisfy Definition~\ref{def} for lower values of $\epsilon_0$ and $\epsilon_1$ are stronger).}
    \label{mix:fig:example}
\end{figure}

We propose to reduce the bias towards getting prematurely stuck in attractors by repeatedly switching the model $M$ used by our sequential scaling methods. If at least one model does not share a given attractor, switching to that model can allow the scaling method to escape it (example in \cref{mix:fig:example}). Furthermore, attractors that contain only incorrect answers may be shared less often because they reflect model-specific biases, whereas correct attractors may be shared more widely because they correspond to solutions intrinsic to the problem. We therefore hypothesise that randomly switching $M$ will help eliminate incorrect attractors while retaining correct ones. This would allow sequential scaling to explore answers beyond the initial basins of attraction of the individual models.

Specifically, we propose an upgrade to any sequential scaling method that replaces the model $M$ with a new model $M_\text{mix}$, which returns the response of a model selected at random from a pool of candidate models $\{M_1, M_2, \ldots\, M_n\}$. In other words, for each sample $Z\sim M_\text{mix}$, we select $j\sim\text{Uniform}(\{1,\ldots,n\})$, and then return $Z\sim M_j(p)$. Using $M_\text{mix}$ changes the models used but does not change the total number of model calls.

\subsection{Evidence that model mixing enables attractor escape}\label{sec:mix:res}

We now show that mixing models makes sequential scaling less likely to get stuck in attractors early on, and more likely to converge to the same attractor across multiple trajectories. We then show that this allows each sequential scaling method to continue to find new answers for longer.

We repeat the experimental setup in \cref{sec:attractors:evidence}, implementing $M_\text{mix}$ on top of the existing test-time scaling baselines. $M_\text{mix}$ selects uniformly from \{GPT OSS 120B, Gemma 3 12B IT, Granite 4.1 8B\}. GPT OSS 120B is the \textit{primary} model (the one with the strongest reasoning capability), while the two non-primary models -- Gemma 3 12B IT and Granite 4.1 8B -- are weaker, to help escape attractors without individually contributing many complete solutions. Appendix~\ref{sec:app:exp} provides further information about our experimental setup.

\begin{table}[t]
\centering
\caption{Comparison of attractor behaviour between scaling methods using $M_\text{mix}$ and methods using only GPT OSS 120B, for the same experiments as in \cref{attractors:tab:part_1}. \textbf{Left columns}: Probability that the scaling method hits an attractor within the first 4 states for a single trajectory. \textbf{Right columns}: Probability that, if two trajectories -- using the same scaling method, model, and question -- both terminate in attractors, then exactly one of the attractors contains only incorrect answers, making it a suboptimal attractor.}
\label{mix:tab:part_2}
\vspace{2pt}
\begin{tabularx}{\textwidth}{llYYY YYY}
\toprule
& &
\multicolumn{3}{c}{Attractor hit by $S_4$ (\%) $\downarrow$} &
\multicolumn{3}{c}{Different correctness (\%) $\downarrow$} \\
\cmidrule(lr){3-5}
\cmidrule(lr){6-8}
Method & Model
& IMO & AMO & RGG
& IMO & AMO & RGG \\
\midrule

\multirow{2}{*}{\parbox{2.8cm}{Reasoning Cache}}
& GPT
& 50.1 & 39.3 & 23.3
& 15.1 & 21.0 & 9.5 \\
& \hl $M_\text{mix}$ (ours)
& \hl{30.3} & \hl{11.1} & \hl{12.7}
& \hl 15.1 & \hl 14.3 & \hl 10.5 \\
\midrule

\multirow{2}{*}{\parbox{2.8cm}{Self-Refine}}
& GPT
& 57.1 & 37.6 & 27.0
& 15.5 & 14.3 & 10.5 \\
& \hl$M_\text{mix}$ (ours)
& \hl{33.4} & \hl{11.1} & \hl{18.7}
& \hl 8.8 & \hl 20.4 & \hl 6.7 \\
\midrule

\multirow{2}{*}{\parbox{2.8cm}{Recursive\\Self-Aggregation}}
& GPT
& 88.7 & 72.6 & 81.0
& 12.7 & 9.8 & 12.7 \\
& \hl$M_\text{mix}$ (ours)
& \hl{67.7} & \hl{51.3} & \hl{49.7}
& \hl 4.7 & \hl 6.3 & \hl 3.3 \\

\bottomrule
\end{tabularx}

\end{table}

\paragraph{Finding 3: Scaling methods with $M_\text{mix}$ are less likely to get stuck in suboptimal attractors.} \Cref{mix:tab:part_2} (left) shows that the sequential methods using $M_\text{mix}$ are less likely to hit attractors within the first four states than the baseline GPT OSS 120B model. This allows them to more often perform additional steps of sequential scaling before progress is halted by an attractor. We demonstrate that these extra steps contribute useful progress in Finding 4. \Cref{mix:tab:part_2} (right) shows that sequential scaling using $M_\text{mix}$ is less likely to find attractors that we can identify as suboptimal.

\begin{figure}
    \vspace{-5pt}
    \centering
    \includegraphics[width=\linewidth]{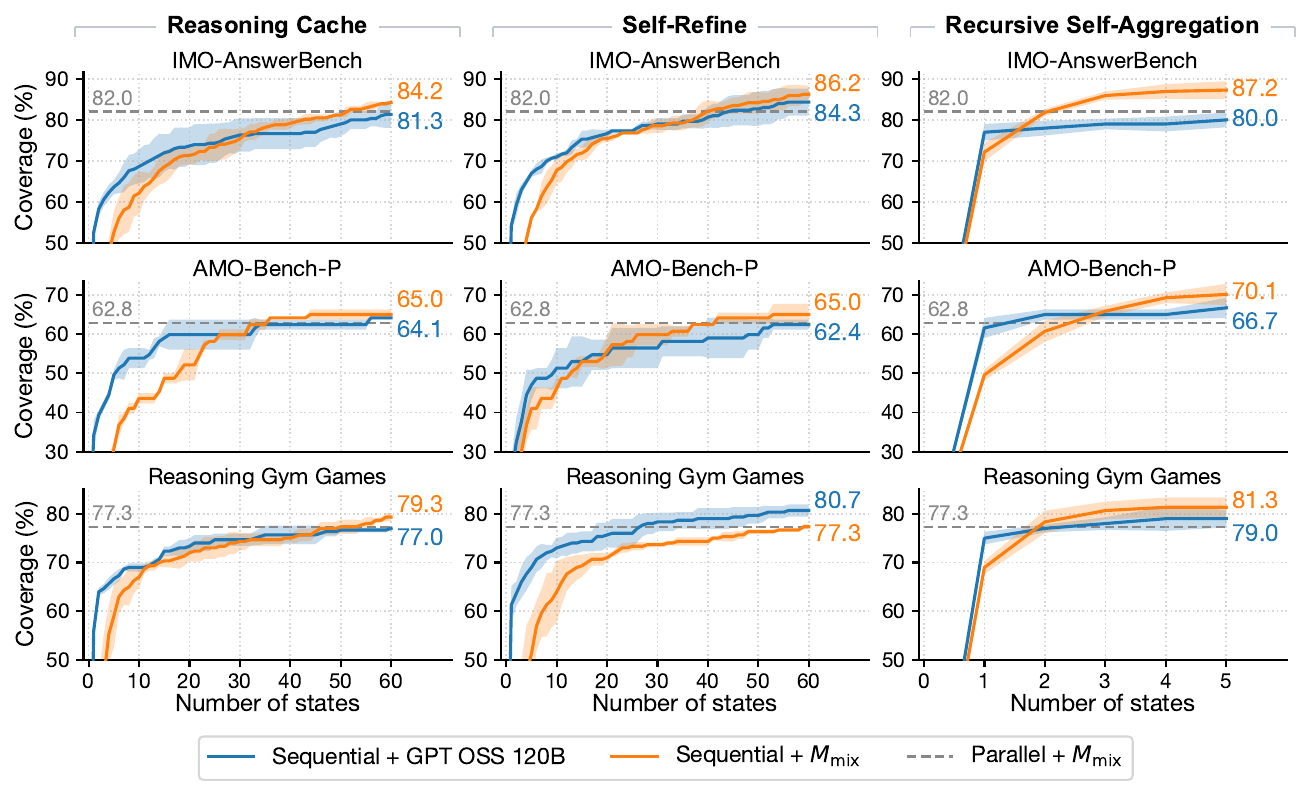}
    \caption{Coverage over time of different sequential methods using $M_\text{mix}$ across three benchmarks. $M_\text{mix}$ (orange) outperforms both the sequential baseline that uses GPT OSS 120B (blue), and the parallel baseline (grey) in almost every case.}
   \label{mix:fig:exploration}
\end{figure}

\paragraph{Finding 4: Sequential methods using $M_\text{mix}$ solve more problems than parallel methods.} To show that a sequential scaling method explores answers that parallel scaling does not, we show that sequential scaling achieves higher \textit{coverage} (the percentage of problems for which we have generated a correct answer \textit{at any stage} of scaling) \citep{brown2024large}. The grey lines in \cref{mix:fig:exploration} show the mean coverage for 60 \textit{parallel} answers from $M_\text{mix}$ (\ie each attempt is conditioned only on the question, such as in self-consistency \citep{wang2023selfconsistency}). While sequential scaling with only the primary model (blue line) could not consistently outperform parallel scaling, sequential scaling using $M_\text{mix}$ (orange line) continues to solve new problems for longer, surpassing both baselines in almost every case. Appendix~\ref{sec:app:example} shows an example of a solution found by only our intervention. We also find that the problems solved by using $M_\text{mix}$ are close to supersets of those from using just the primary model (Appendix~\ref{sec:app:conf}). We find similar effects on coverage when comparing to a wide range of baselines, including increasing temperature as an alternative method for improving diversity, and compute-matching by cost rather than number of states (Appendices~\ref{sec:app:cost}-\ref{sec:app:temperature}).

\begin{figure}[t]
    \vspace{-3pt}
    \centering
    \includegraphics[width=\linewidth]{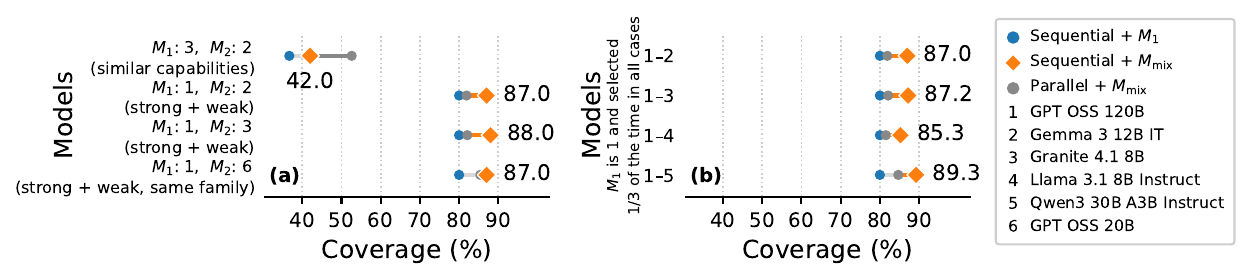}
    \vspace{-10px}
    \caption{Coverage for alternative model mixtures for RSA on IMO-AnswerBench. Left: Results for different combinations of models, to identify whether some models are better at escaping attractors than others. Right: Results for different numbers of non-primary models where the primary model is always selected with probability $\nicefrac{1}{3}$, to test whether adding more models provides meaningfully more opportunities for one model being able to escape attractors.}\label{mix:fig:ablations}
    \vspace{-8pt}
\end{figure}

\paragraph{Finding 5: Model mixing is effective across different combinations of models.} \Cref{mix:fig:ablations} shows that sequential methods using $M_\text{mix}$ achieve better coverage than those using a single strong model across different candidate model pools (additional omitted experiments are in Appendix~\ref{sec:app:fail}). The improvements are weakest using models from the same family (GPT OSS 20B and 120B), or of similar capabilities (Gemma 3 12B IT and Granite 4.1 8B), while all other configurations produce relatively uniform improvements. We further show that our results transfer to larger models by testing the same configuration with Qwen3 235B A22B Thinking 2507 \citep{qwen3technicalreport} and GPT OSS 120B (both medium reasoning effort), where Qwen is selected with probability 0.75.

\begin{figure}[t]
    \centering
    \includegraphics[width=0.85\linewidth]{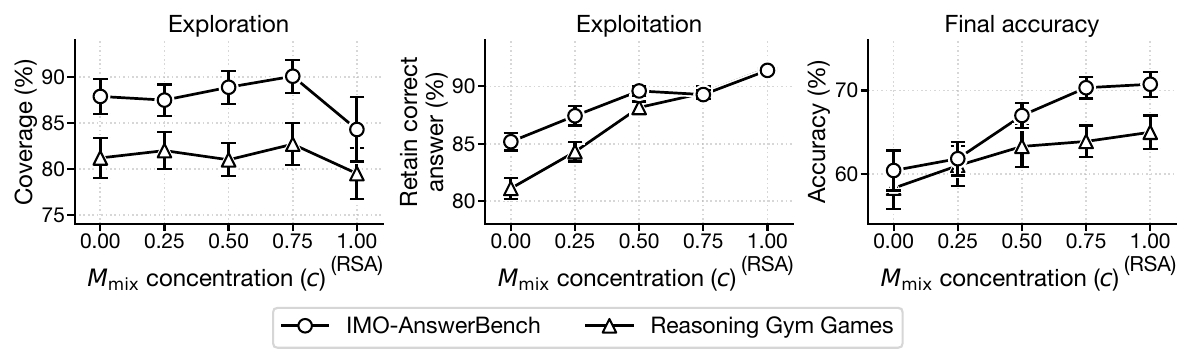}
    \vspace{2pt}
    \caption{Effect of mixture concentration $c$ on RSA with $M_\text{mix}$ over 10 states. \textbf{Left}: Coverage is maximised by occasionally sampling weaker models. \textbf{Centre}: Probability of generating a correct answer when there was one in the previous state improves as sampling concentrates on the primary model. \textbf{Right}: Using a single model outperforms mixtures of models overall.}
    \label{mix:fig:number}
    \vspace{-13pt}
\end{figure}

\paragraph{Finding 6: Model mixing is most effective when the primary model is used most often.} We now test how the probability of selecting the primary model (the model with the strongest reasoning capability) changes the performance. Let us assume $M_\text{mix}$ selects from a set of three models $\{M_1, M_2, M_3\}$ with $M_1$ being the primary model. We parametrise $M_\text{mix}$ by $c\in[0,1]$ so that we sample these three models with probabilities $\frac{1}{3}\times[1+2c,1-c,1-c]$ ($0\le c\le1$) respectively. For example, $c=0$ would yield uniform sampling, and $c=1$ would yield sampling only $M_1$ (\ie no model mixing). \Cref{mix:fig:number} (left) shows that the best exploration comes from $c=0.75$. This may be because non-primary models are useful in order to escape attractors, but sampling them any more than the minimum just slows progress by overusing weak models. Conversely, \Cref{mix:fig:number} (centre) shows that the default RSA setup with just the primary model is the best at retaining correct answers. This also achieves the highest final accuracy (probability of $Y_T$ being correct).

\paragraph{Finding 7: Mixing models can improve final accuracy.} \Cref{mix:fig:number} shows that simple mixtures can provide strong exploration, but fail to improve over using single models because they are also more likely to escape \textit{correct} answers. This could be true even given our hypothesis that correct answers are more likely to remain as attractors for the mixture. To help demonstrate that mixing models can also improve overall performance, Appendix~\ref{sec:app:ams} presents a simple annealing method for balancing increased exploration from model mixing with stronger correct-answer exploitation from using just the primary model. Our annealed model sampling method improves accuracy for RSA over using just GPT OSS 120B, by 2.2 percentage points on IMO-AnswerBench, and 2.3 points on Reasoning Gym Games, and at a lower cost.

\begin{takeaway}
\vspace{-2pt}
\begin{points}
  \point{i}{Mixing models makes sequential scaling less likely to get prematurely stuck in attractors;}
  \point{ii}{This allows sequential scaling to access problems that parallel scaling does not;}
  \point{iii}{These findings show the potential of sequential scaling methods over long timescales.}
\end{points}
\end{takeaway}

\section{Conclusion}

In this work, we show that, across 27 combinations of sequential scaling methods, models, and benchmarks, sequential scaling often fails to explore because it becomes trapped in suboptimal attractors, but that a simple model-mixing intervention helps escape these attractors and access solutions beyond parallel scaling.

We highlight two main limitations in our experiments due to cost constraints. First, we test a limited range of configurations. For example, further investigation is necessary to confirm for which tasks attractors do not appear (see Appendix~\ref{sec:app:arc}). Second, we do not extend our investigation to frontier models.

Our results motivate refocusing long-horizon test-time scaling from parallel methods (a search process under a fixed model policy) to sequential methods that improve on previous answers (a learning process where the model iteratively builds on its ideas to improve its search). Future work could investigate whether better attractors can be found by doing sequential scaling with alternative state representations \citep{dehghani2018universal,phan2025migrate}, or optimising components of sequential scaling methods \citep{zhou2026multi,agrawal2026gepa,qu2024recursive} against the coverage metric.

\section*{Acknowledgements}

JR is supported by an Industrial Doctoral Landscape Award studentship funded by the Engineering and Physical Sciences Research Council and BBC Research \& Development. PT and AB acknowledge support from the UKRI Turing AI Fellowship (EP/W002981/1). AB also acknowledges support from Coefficient Giving.

\section*{Reproducibility Statement}

Appendix~\ref{sec:app:exp} provides all the details required to reproduce our results, including the experimental setup, prompts, model configurations, and evaluation procedures. A link to the code we use to run our experiments is included at the end of the abstract.

\section*{AI Use Statement}

Outside of the models studied in our experiments, language models were used in the following ways in the writing of this paper:

\begin{itemize}
    \item To help search for relevant prior work, for example based on a short explanation of the paper.
    \item To help improve efficiency of experiments, for example by adding parallelism to the code. All of our experiments were initially written by hand, and the changes made by language models did not change the underlying operations performed in the experiments.
    \item To make existing figures more aesthetically pleasing, for example to change the colours.
    \item To help critique and improve on individual sentences of this paper, without generating full paragraphs at once.
    \item Language models were not used for research ideation.
\end{itemize}

We take responsibility for the final content of this work.

\bibliography{iclr2027_conference}
\bibliographystyle{iclr2027_conference}

\appendix

\section{Scope of Test-time Scaling Methods Investigated}\label{sec:app:scope}

We focus only on test-time scaling methods that operate entirely through natural language using off-the-shelf models without task-specific feedback. Below are some popular methods not in the scope of our investigation.

\begin{itemize}
    \item Methods that scale in a learnt latent space that is updated on each forward pass \citep[eg][]{dehghani2018universal}. \cite{huang2026} provide similar analysis for these methods.
    \item Methods that do gradient updates to the model at test-time \citep{phan2025migrate, yuksekgonul2026learning, hubert2026olympiad}, because these methods are often costly in this regime as they must compress their full experience into their weights at once while also maintaining pretrained knowledge.
    \item Methods similar to vanilla chain-of-thought \citep{wei2022chain, goyal2024think}, because they can only be scaled up to the model's context length. We therefore treat a long chain-of-thought as a single model call.
    \item Methods that rely on problem-specific tools such as a REPL, or a verifier that is fine-tuned for the current task \citep{lightman2024lets, snell2024scaling, romera2024mathematical, zheng2026dream}.
    \item Methods that use subagents \citep[eg][]{zhang2025recursive} because they are primarily specialised for problems that can be decomposed into subproblems.
\end{itemize}

\section{Experimental Setup}\label{sec:app:exp}

We repeat every experiment across three random seeds and present the mean, except for in \cref{mix:fig:ablations,app:fig:temperature,app:fig:fail}, where we used only one seed to reduce costs.

\subsection{Benchmarks}

We focus our experiments around the following three benchmarks.

\begin{itemize}
    \item \textbf{IMO-AnswerBench (IMO).} 400 difficult maths questions selected from past mathematical olympiad competitions \citep{luong-etal-2025-towards}.
    \item \textbf{AMO-Bench-P (AMO).} A smaller set of 39 maths questions \citep{an2025amobench}, which are less saturated than those of IMO-AnswerBench.
    \item \textbf{Reasoning Gym Games (RGG).} A set of 100 problems used by \cite{venkatraman2025recursive}, originally generated from the Reasoning Gym benchmark \citep{stojanovski2026reasoning}. These problems cover a range of small games, such as finding the best move in a simplified game of Go.
\end{itemize}

In all three cases, the benchmark defines a set of problems, each consisting of a text-based question, and a single correct answer.

\subsection{Scaling methods}

We test the three sequential scaling methods that are described in \cref{back:tab:methods}. In \cref{attractors:tab:part_1} we test the methods with each of the following models: GPT OSS 120B (medium reasoning effort) \citep{openai2025gptoss}, Gemma 3 12B IT \citep{gemmateam2025gemma3}, Granite 4.1 8B \citep{granite2026}. We select these models to cover a diverse set of model families and capabilities, while also aligning with our experiments in \cref{sec:mix}, where we test \textit{identical} sequential scaling setups, but with $M_\text{mix}$ as the model instead. We sample with a temperature of 1 (except in \cref{app:fig:temperature}) for all models except for Granite 4.1 8B, where we use 0.2.

We use the same three models for our main experiments with $M_\text{mix}$, where GPT OSS 120B is the \textit{primary} model (the one with the strongest reasoning capability). The two non-primary models -- Gemma 3 12B IT and Granite 4.1 8B -- are included to help escape attractors. We select weaker non-primary models so that they are unlikely to individually contribute complete solutions to problems the primary model cannot solve, to help isolate their effect on the dynamical system.

In our experiments, we generate $T=60$ states per trajectory for reasoning cache and self-refine, and $T=5$ states for recursive self-aggregation. We use $N=12$ answers per state in recursive self-aggregation, and use $N=60$ individual answers for our parallel baselines, so that all methods are matched in their total number of attempts to solve the problem. In \cref{intro:fig:main,mix:fig:number,mix:fig:final}, these $T$ values are doubled, to demonstrate capabilities over long timescales. We more directly compare compute cost in Appendix~\ref{sec:app:cost}.

\subsubsection{Reasoning cache}

We use the same prompts in our reasoning cache implementation as \citet{wu2026reasoning}, with minor changes made to remove the references to `maths' for the Reasoning Gym Games benchmark, and with the summary length set to `one-page'. Unless stated otherwise, we generate 60 states (\ie pairs of a summary and an answer) for each experiment.

\subsubsection{Self-refine}

We adapt the prompts presented by \citet{madaan2023self} so that they apply across a range of tasks. Specifically, we use the following system prompt when generating the critique.

\begin{promptplain}
You are a strict grader. Review the solution provided. Identify any logical errors, missing justifications, calculation mistakes, or unclear reasoning. Provide constructive feedback focusing on what is incorrect or could be improved.
\end{promptplain}

We use the following system prompt for refining the previous answer based on a critique.

\begin{promptplain}
Based on the feedback provided, refine and improve your solution. Ensure that all errors are corrected, gaps are filled, and the logic is rigorous. End with your final answer in \boxed{}.
\end{promptplain}

Unless stated otherwise, we generate 60 states (\ie answers conditioned on a critique) for each experiment.

\subsubsection{Recursive self-aggregation}

We use the same prompts in our recursive self-aggregation implementation as \citet{venkatraman2025recursive}. Unless stated otherwise, we generate $T=5$ states where each state consists of $N=12$ answers. This is primarily because generating 60 states would be prohibitively expensive, and there is no need to match compute across sequential scaling methods as we do not compare these different methods to each other. In every case for recursive self-aggregation we aggregate $K=3$ answers from the previous state. We select these parameters as the lowest cost configuration that achieves close to the best performance in the experiments by \citet{venkatraman2025recursive}.

\subsubsection{Parallel scaling}

For experiments that measure coverage, we compare to a parallel scaling baseline consisting of $N$ independent answers from the language model (\eg as in \citet{wang2023selfconsistency}). Because we measure coverage, we do not need to implement a method to select between these $N$ answers. For the independent answers, the language model is given only the question in its prompt, with a simple system prompt to account for very similar prompts appearing in the test-time scaling methods:

\begin{promptplain}
You are given a question. Reason step-by-step and return your final answer in \boxed{}.
\end{promptplain}

To obtain the parallel scaling scores in Figures \ref{mix:fig:exploration} and \ref{mix:fig:ablations}, we generate 180 independent answers for each question from the relevant model, and compute the number of problems solved at least once when sampling $N=60$ random answers from the 180. We repeat this 50 times and report the mean. The standard deviation in coverage between three disjoint sets of these 60 random answers is too low to display on any plots.

\subsection{Grading}

Answers generated by a language model generally consist of a \textit{final answer}, for example a single number or mathematical expression, surrounded by an explanation for why this answer is correct. In all three benchmarks, only the final answer is checked against the golden answer provided by the benchmark.

For AMO-Bench-P, Reasoning Gym Games, and ARC AGI 1, we use the grading logic proposed in the respective papers \citep{an2025amobench, stojanovski2026reasoning, chollet2019measure}. For IMO-AnswerBench, we use the same grader prompts as \citet{luong-etal-2025-towards} to extract and compare final answers, but using GPT OSS 20B (low reasoning effort) instead of Gemini 2.5 Pro, to save on compute costs. We find that the use of a weaker grader model yields little loss of grader quality, and does not increase the uncertainty in our results enough to change our conclusions. Specifically, GPT OSS 20B generates identical grades to Gemini 2.5 Pro on 99.25\% of answers across 2,000 random samples from our experiments. We find no evidence that grading disagreements are systematically associated with a particular scaling method, and use Gemini to test the problems that change correctness when using $M_\text{mix}$ in \cref{app:fig:conf} to find only 17 grading disagreements across all problems, which did not lead to any changes to the overall figures.

\subsection{Attractors}

\subsubsection{Identifying attractors from text data}

\begin{algorithm}[t]
\caption{Identifying attractors from sampled trajectories}
\label{alg:attractors}
\begin{algorithmic}[1]
\Require Summariser $M_{\mathrm{sum}}$ (GPT OSS 20B);
         embedder $M_{\mathrm{emb}}$ (Gemini Embedding~2, preview)
\Statex
\Function{Embed}{$X, Y$}
  \Comment{problem $X$, answer to embed $Y$}
  \State $\textit{sPrompt} \gets \textsc{SummaryPrompt}(X, Y)$
  \State $\textit{summary} \gets M_{\mathrm{sum}}\!\left(\textit{sPrompt};\ \text{temperature} = 1,\ \text{reasoning effort} = \textsc{low}\right)$
  \State $\textit{ePrompt} \gets \textsc{QueryPrompt}(\textit{summary})$
  \State $v \gets M_{\mathrm{emb}}\!\left(\textit{ePrompt};\ \text{output dimension} = 256\right)$
  \State \Return $v / \lVert v \rVert_2$
    \Comment{unit vector in $\mathbb{R}^{256}$}
\EndFunction
\Statex
\Function{GetAttractors}{$X$, $A$}
  \Comment{$A$: states from multiple seeds with the same config}
  \State $E \gets \left\{\, \Call{Embed}{X, Y} \;:\; Y \in A \,\right\}$
  \State $C \gets{}$ \Call{KMeans}{$E$, num clusters $= \lceil |E| / 12 \rceil$} \Comment{one centroid per cluster}
  \State \Return $\Call{Attractors}{C,\ \epsilon_0 = 0.1,\ \epsilon_1 = 0.1}$ \Comment{Definition~\ref{def:attractor}, by exhaustive search}
\EndFunction
\end{algorithmic}
\hrule
\small
\vspace{8pt}

\textsc{SummaryPrompt}$(X, Y)$:
\vspace{-5pt}
\begin{promptplain}
Write a four-sentence, technical summary of the main points of the
below solution. Do not include the problem in your summary. Try to
state the names of theorems or constructions you use.

Problem:
{X}

Solution to summarise:
{Y}
\end{promptplain}
\vspace{-1pt}
\textsc{QueryPrompt}$(\textit{summary})$:
\vspace{-5pt}
\begin{promptplain}
task: theorem similarity | query: {summary}
\end{promptplain}
\vspace{-6pt}
\end{algorithm}

We identify attractors following Definition~\ref{def}, from combinations of three independently-sampled trajectories with identical configurations. We augment the attractor definition by requiring
\begin{itemize}
    \item Each attractor must have at least 10 states to avoid satisfying the $\epsilon_0$ bound by counting a very small number of transitions.
    \item Attractors must cover fewer than 60\% of clusters to avoid the degenerate case where an attractor covers most states across different seeds, as such an `attractor' necessarily almost always satisfies the entry and retention probability requirements, but is rarely interesting because it is not constrained to a specific part of the potential answer space.
\end{itemize}
These values must be selected to avoid degenerate cases while minimising other effects. To ensure this, we selected these values \textit{before} viewing the results in \cref{attractors:tab:part_1} to avoid biasing our choices, and verify that they lead to meaningful attractors in \cref{sec:app:verif}.

Because answers that contain different tokens but are mathematically equivalent should be considered to belong to the same state, we group together mathematically equivalent answers, following \cref{alg:attractors} \citep{prinz2011markov}. We select the configuration in \cref{alg:attractors} to find, as an objective heuristic, the largest clusters that ensure a low probability of a cluster containing states that result in different final answers. We then apply Definition~\ref{def} by exhaustive search over a pruned tree of possible sets of clusters, using the combined transitions from three seeds with the same configuration. \Cref{sec:app:verif} shows that the configuration we selected finds meaningful attractors.

\subsubsection{Identifying attractors in recursive self-aggregation}

For recursive self-aggregation, each state contains 12 answers, so generating enough answers to directly estimate transitions between complete states would be prohibitively expensive. We instead identify attractors from transitions between individual answers and use these to assign attractor membership to complete states. In particular, we record a transition whenever one answer was included in the context used to generate another, apply the clustering procedure in \cref{alg:attractors} to the answers, and identify attractors in the resulting cluster transition graph. For each answer-level attractor $A^{(\text{ans})}$ we estimate the retention requirement for attractors as the fraction of generated answers that belong to $A^{(\text{ans})}$, out of those with a context containing at least one answer from $A^{(\text{ans})}$, to avoid double counting instances that conditioned on multiple answers from $A^{(\text{ans})}$ at once. We use an answer-level escape threshold of $\epsilon_0^{\text{(\text{ans})}}=0.2985$, so that $\epsilon_0$ will become 0.1 after we transform it to the state level. We assign a state to state-level attractor $A$, when at least six of its twelve answers belong to the corresponding answer-level attractor $A^{(\text{ans})}$.

The resulting state-level attractors satisfy the retention and basin-entry conditions of Definition~\ref{def} for $\epsilon_0=\epsilon_1=0.1$, while allowing us to identify these attractors through answer-level transition dynamics. Specifically, for a particular answer-level attractor $A^{(\text{ans})}$ a new answer belongs to $A^{(\text{ans})}$ with probability at least $1-\epsilon_0^{(\text{ans})}=0.7015$, when its context includes at least one answer from $A^{(\text{ans})}$. Here, we assume that the distribution of the other two answers in context is always consistent, which could be false, but is a reasonable assumption in practice (see Appendix~\ref{sec:app:verif}). Therefore, for any current state containing at least six answers in $A^{(\text{ans})}$, the probability that a given answer in the next state belongs to $A^{(\text{ans})}$ is at least
$$
P\left(S_{t+1}^i\in A^{(\text{ans})}\middle|\sum_{j=1}^{12} \mathbbm{1}_{S_t^j\in A^{(ans)}}\ge 6\right) \ge 0.7015\left(1-\frac{6}{12}\frac{5}{11}\frac{4}{10}\right)\approx0.6377,
$$
which gives the required $\epsilon_0$ bound at the state-level. A similar bound can be obtained for $\epsilon_1$ by considering states containing at least five answers already in the attractor, followed by $k=2$ transitions.

\subsubsection{Validating our attractor definition}\label{sec:app:verif}

\begin{table}[t]
\centering
\caption{Probability that a state in an attractor transitions to a next state in the same attractor. For each combination of model, scaling method, and benchmark, we randomly sample 100 states that are in attractors. We resample a single next state from each one, and record how often this next state lies within the same attractor. We do not recompute any clusters, and assign points to the cluster with the closest centroid. For RSA, a state is considered to be in an attractor if it contains at least 6 states from that attractor. Our results show that the next state is in the same attractor with probability $>1-\epsilon=0.9$ in almost every case.}
\label{app:tab:cont}
\vspace{2pt}
\begin{tabularx}{0.75\textwidth}{llYYY}
\toprule
& &
\multicolumn{3}{c}{Same attractor (\%) $\uparrow$} \\
\cmidrule(lr){3-5}
Method & Model
& IMO & AMO & RGG \\
\midrule

\multirow{3}{*}{\parbox{2.8cm}{Reasoning Cache}}
& GPT OSS 120B
& 96 & 99 & 98 \\
& Gemma 3 12B IT
& 100 & 96 & 97 \\
& Granite 4.1 8B
& 94 & 96 & 96 \\

\midrule

\multirow{3}{*}{\parbox{2.8cm}{Self-Refine}}
& GPT OSS 120B
& 96 & 100 & 98 \\
& Gemma 3 12B IT
& 95 & 94 & 99 \\
& Granite 4.1 8B
& 93 & 92 & 96 \\

\midrule

\multirow{3}{*}{\parbox{2.8cm}{Recursive\\Self-Aggregation}}
& GPT OSS 120B
& 95 & 100 & 98 \\
& Gemma 3 12B IT
& 100 & 97 & 95 \\
& Granite 4.1 8B
& 96 & 97 & 97 \\

\bottomrule
\end{tabularx}

\end{table}

\begin{figure}
    \centering
    \includegraphics[width=0.95\linewidth]{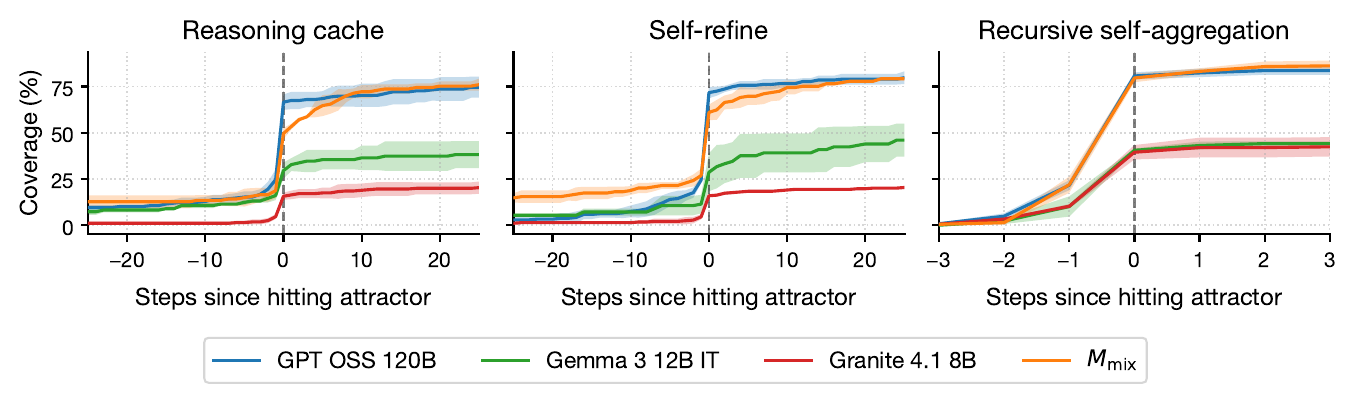}
    \caption{Coverage before and after entering an attractor on IMO-AnswerBench, with trajectories aligned to the state where they enter the attractor (grey dashed line). Progress slows substantially after entry, indicating that we identify attractor states that limit further exploration.}
    \label{app:fig:proof}
\end{figure}

In the previous section we outline how we identify attractors from trajectories. Our goal is to identify attractors that follow Definition~\ref{def} in practice: once entered, an attractor should prevent transitions to states that are semantically different to the centroids of its constituent clusters. \Cref{app:tab:cont} shows that resampling \textit{new} transitions from states in attractors leads to the next state being in the same attractor with probability $>1-\epsilon=0.9$ in almost every case. Furthermore, \cref{app:fig:proof} demonstrates that, after we hit an attractor (vertical grey line), progress towards correct answers slows to a halt almost immediately after entering an attractor (the gradients of all lines flatten after passing the vertical dashed line). In \cref{app:fig:proof}, the large jump before entering an attractor is likely due to attractors appearing once we hit correct answers. This does not necessarily mean the state transitions prior to entering the attractor are not important: they may have played a useful role in approaching the correct answer.\footnote{For example, let $P_t(a)$ be the probability that the sequential scaling method emits an answer $a$ within the first $t$ states, and $P_t(a\mid s)$ the probability starting from state $s$. Then, we could define the value of a state as: $V_t(s\mid X) \;=\; \sup_{a:\,C_X(a)=1}\Big[\log P_t(a\mid s) - \log P_t(a)\Big]$, where $C_X(a)$ is true iff $a$ solves $X$ - \ie the amount of useful information that $s$ supplies our model \citep{finzi2026entropy}. Under this formulation, the value of states would increase more steadily as they approach a correct answer than what appears in the graph.}

\begin{table}[t]
\centering
\caption{Sensitivity of attractor hitting-time probabilities to randomness in
\cref{alg:attractors} for IMO-AnswerBench (\ie standard deviation for left column of \cref{attractors:tab:part_1}). Results are stable across two different seeds used in
\cref{alg:attractors}.}
\label{app:tab:proof}

\begin{tabular}{llc}
\toprule
Method & Model & Attractor hit by $S_4$ (\%) $\downarrow$ \\
\midrule

\multirow{3}{*}{\parbox{2.8cm}{Reasoning Cache}}
& GPT OSS 120B
& $47.67 \pm 2.33$ \\
& Gemma 3 12B IT
& $46.67 \pm 1.33$ \\
& Granite 4.1 8B
& $73.50 \pm 0.50$ \\

\midrule

\multirow{3}{*}{\parbox{2.8cm}{Self-Refine}}
& GPT OSS 120B
& $57.31 \pm 0.33$ \\
& Gemma 3 12B IT
& $44.33 \pm 3.67$ \\
& Granite 4.1 8B
& $69.17 \pm 2.17$ \\

\midrule

\multirow{3}{*}{\parbox{2.8cm}{Recursive\\Self-Aggregation}}
& GPT OSS 120B
& $87.17 \pm 2.83$ \\
& Gemma 3 12B IT
& $59.83 \pm 3.83$ \\
& Granite 4.1 8B
& $69.50 \pm 2.17$ \\

\bottomrule
\end{tabular}

\end{table}

We show that these results are stable by recomputing the attractors with a different seed (which affects summaries, embeddings, and clusters). \Cref{app:tab:proof} shows the mean and standard deviation of the first column of \cref{attractors:tab:part_1} after we recomputed the attractors a single time for a random set of 100 questions. \Cref{app:tab:proof} shows that the hit rates of attractors we identify are unlikely to change much under different random seeds.

\subsubsection{Example of an attractor we identify}

We now provide an example of how attractors are identified, using two trajectories generated using self-refine for the following IMO-AnswerBench problem. We present these trajectories because they have the common features of attractors that we wish to highlight.

\begin{benchproblem}[title={IMO-AnswerBench}]
What is the least positive integer $n$ such that, in every convex 1001-sided polygon, the sum of any $n$ diagonals is greater than or equal to the sum of the remaining diagonals?
\end{benchproblem}

\begin{figure}[t]
    \centering
    \includegraphics[width=0.7\linewidth]{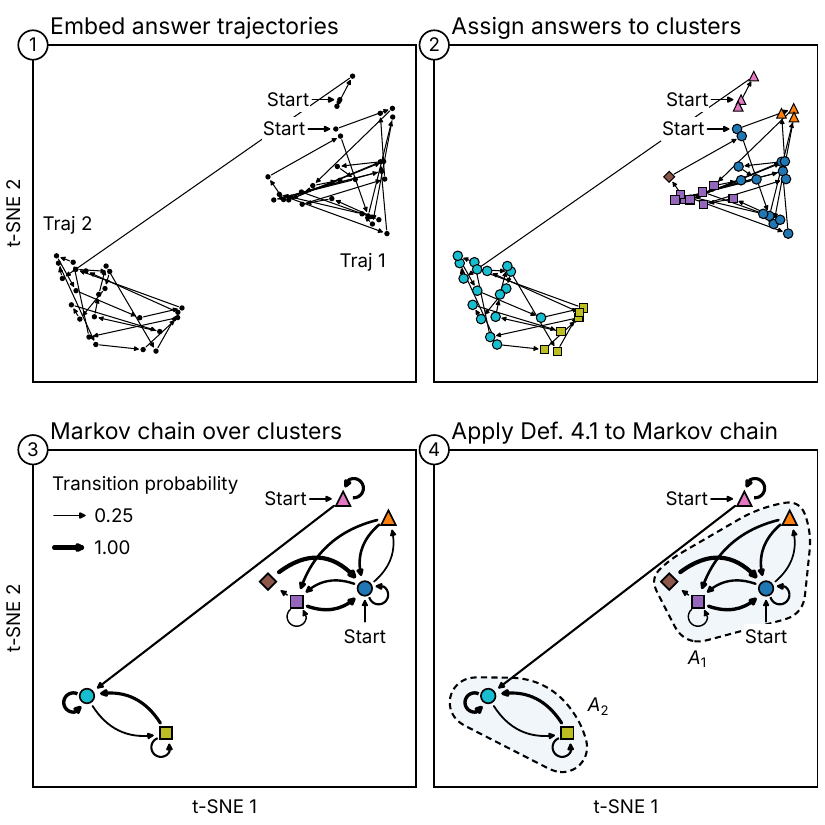}
    \caption{Visual representation of the process of finding attractors in the first 25 states from 2 partial trajectories generated by self-refine on a problem from IMO-AnswerBench.}
    \label{app:fig:proc}
\end{figure}

\Cref{app:fig:proc} displays the process of finding attractors in first 25 states of each trajectory (in our results, we record 3 different trajectories of 60 states each, but we omit the full set of states here improve clarity). Attractors are identified from trajectories of states through the below four step process, based on the \verb|GetAttractors| function in \cref{alg:attractors}, (with line 11 of \cref{alg:attractors} combining steps \circled{3} and \circled{4}).

\begin{itemize}
    \item[\circled{1}] \textbf{Embed answer trajectories.} We embed the text from each state $S_i$ into a 256-dimensional vector, using the \verb|Embed| function in \cref{alg:attractors}. \Cref{app:fig:proc} displays a 2D projection of this vector obtained using t-SNE for display only, with arrows between each $S_i$ and $S_i+1$. In the case of \cref{app:fig:proc}, $S_i$ is simply the text of the most recent answer.
    \item[\circled{2}] \textbf{Assign answers to clusters.} We use K-means to assign each answer to a cluster.
    \item[\circled{3}] \textbf{Construct a Markov chain over clusters.} The states of the Markov chain are the clusters we identify in step \circled{2}, and the transition probability between a pair of clusters $a$ and $b$ is the probability that a randomly sampled answer in cluster $a$ transitions to a next state in cluster $b$.
    \item[\circled{4}] \textbf{Apply Definition~\ref{def} to the Markov chain to identify sets of clusters that are attractors.} As a special case that can be seen in $A_1$ in our example, we consider any set of clusters that contains the starting state to satisfy condition ii of Definition~\ref{def}.
\end{itemize}

In \cref{app:fig:proc}, Traj 1 begins in an attractor, as it repeatedly generates its \textit{correct} initial answer answer of 499,000. In contrast, Traj 2 never generates this answer, and becomes stuck in a attractor when it arrives at $S_4$. The attractor of Traj 1 is optimal because it contains only correct answers, while Traj 2 finds a suboptimal attractor that does not contain any correct answers.

Traj 2 fails to reach a correct answer because it gets stuck in this suboptimal attractor rather than exploring new ideas. Each of its first 3 states contains a trivial error: $S_1$ claims that any $n$ larger than half the number of diagonals satisfies the requirement, which is trivially false; $S_2$ correctly identifies that it must minimise $n$ in the worst case, but incorrectly uses the $n$ \textit{longest} diagonals, which would be the best case; and $S_3$ corrects the reasoning to test the $n$ shortest diagonals, but writes `a natural choice is to select half of the diagonals plus one' and provides the same final answer as the first two states. From $S_4$ onwards, progress halts as all answers follow the same logic, which attempts to correct the answer from $S_3$ through a long (and false) explanation of why a regular 1001-sided polygon must be the worst case scenario.

\subsubsection{Attractor metrics}\label{sec:app:met}

\begin{figure}[t]
    \centering
    \includegraphics[width=0.9\linewidth]{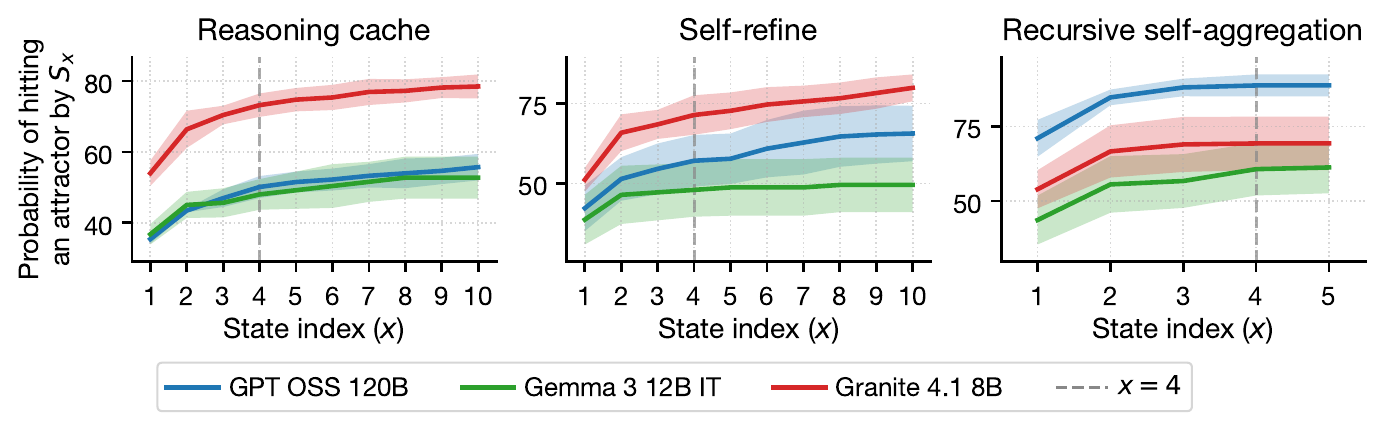}
    \caption{Probability of entering an attractor at or before state index $x$ for IMO-AnswerBench. The probability begins to level off after four states, motivating the four-state threshold used throughout our experiments.}
    \label{app:fig:steps}
\end{figure}

In sections \ref{sec:attractors:evidence} and \ref{sec:mix:res}, we present the probability that each sequential scaling method hits an attractor within the first four states. We select this threshold, because, if we set the threshold earlier, it would not yet provide evidence that the hit rate eventually becomes high, while if we use a later threshold, it would only provide the weaker conclusion that this is true at a later state (see \cref{app:fig:steps}).

\section{Probability That Different Trajectories Hit Different Attractors}\label{sec:app:corr}

In \cref{attractors:tab:part_1}, we measure the probability that, if we generate two trajectories using the same sequential scaling method and model, on the same question, and both terminate in attractors, then exactly one of the attractors contains only incorrect
answers. To show the presence of even more attractors in the answer space with potentially varying correctness, we now measure how often two trajectories reach different attractors regardless of correctness (e.g. two attractors that both contain only incorrect states, but have no states in common). \Cref{attractors:tab:corr} shows that it is common for the same scaling-method--model combination to find different attractors when run with different random seeds. The existence of multiple attractors implies the answer space may have even more suboptimal attractors than are identified in \cref{attractors:tab:part_1}. The values in \cref{attractors:tab:corr} reduce for the scaling methods using $M_\text{mix}$, implying the starting distribution $f_s(X,M_\text{mix})$ can be partitioned into fewer basins of attraction on average, potentially meaning there are fewer suboptimal attractors that we could get stuck in on our way to a better attractor.

\begin{table}[t]
\centering
\caption{Probability that, if we generate two trajectories using the same scaling method, model, and question, and both terminate in attractors, then the two attractors are different. Higher values suggest there are more attractors in the state space.}
\label{attractors:tab:corr}
\vspace{2pt}
\begin{tabularx}{0.75\textwidth}{llYYY}
\toprule
& &
\multicolumn{3}{c}{Different attractors (\%) $\downarrow$} \\
\cmidrule(lr){3-5}
Method & Model
& IMO & AMO & RGG \\
\midrule

\multirow{3}{*}{\parbox{2.8cm}{Reasoning Cache}}
& GPT OSS 120B
& 33.7 & 22.2 & 31.0 \\
& Gemma 3 12B IT
& 34.3 & 18.9 & 32.2 \\
& Granite 4.1 8B
& 68.0 & 65.3 & 63.7 \\
& \hl $M_\text{mix}$ (ours)
& \hl{20.5} & \hl{20.0} & \hl{8.2} \\

\midrule

\multirow{3}{*}{\parbox{2.8cm}{Self-Refine}}
& GPT OSS 120B
& 47.3 & 47.6 & 23.7 \\
& Gemma 3 12B IT
& 43.8 & 43.6 & 20.2 \\
& Granite 4.1 8B
& 74.5 & 79.3 & 40.6 \\
& \hl $M_\text{mix}$ (ours)
& \hl{25.0} & \hl{33.3} & \hl{20.0} \\

\midrule

\multirow{3}{*}{\parbox{2.8cm}{Recursive\\Self-Aggregation}}
& GPT OSS 120B
& 12.4 & 10.8 & 12.0 \\
& Gemma 3 12B IT
& 4.9 & 8.7 & 7.7 \\
& Granite 4.1 8B
& 8.8 & 9.3 & 11.3 \\
& \hl $M_\text{mix}$ (ours)
& \hl{5.9} & \hl{4.3} & \hl{1.9} \\

\bottomrule
\end{tabularx}

\end{table}
\section{Example Solution Only Identified by a Sequential Scaling Method with $M_\text{mix}$}\label{sec:app:example}

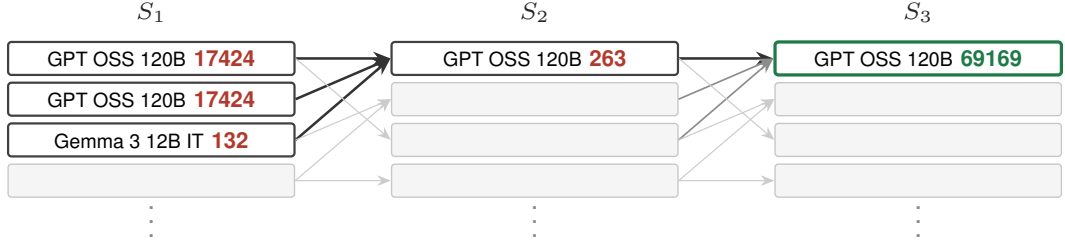
\begin{figure}[t]
\centering
\resizebox{\textwidth}{!}{%
\begin{tikzpicture}[
  font=\sffamily\small, x=1cm, y=1cm,
  slot/.style={
    rounded corners=1pt, draw=black!75, fill=white, line width=0.65pt,
    inner xsep=4pt, inner ysep=1.5pt, align=center,
    minimum width=2.95cm, minimum height=0.34cm},
  final/.style={slot, draw=rightanswer, line width=0.95pt},
  empty/.style={slot, draw=black!22, fill=black!4, line width=0.4pt},
  flow/.style={-{Stealth[length=1.8mm,width=1.4mm]}, draw=black!80, line width=0.75pt},
  contrib/.style={-{Stealth[length=1.6mm,width=1.25mm]}, draw=black!45, line width=0.5pt},
  faint/.style={-{Stealth[length=1.4mm,width=1.1mm]}, draw=black!20, line width=0.35pt},
  hdr/.style={font=\sffamily\scriptsize\bfseries, text=black!85, anchor=south},
]

\newcommand{\mdl}[1]{{\fontsize{6.4}{7}\selectfont #1}}
\newcommand{\ansW}[1]{{\fontsize{7.2}{7}\selectfont\textbf{\textcolor{wronganswer}{#1}}}}
\newcommand{\ansR}[1]{{\fontsize{7.2}{7}\selectfont\textbf{\textcolor{rightanswer}{#1}}}}

\def\cA{0} \def\cB{3.95} \def\cC{7.90}
\def\dy{0.42}

\foreach \i in {0,...,3}{
  \node[empty] (pa\i) at (\cA, {0.90-\dy*\i}) {};
  \node[empty] (pb\i) at (\cB, {0.90-\dy*\i}) {};
  \node[empty] (pc\i) at (\cC, {0.90-\dy*\i}) {};
}
\foreach \x in {\cA,\cB,\cC}{ \node[text=black!45] at (\x, {0.90-\dy*3.75}) {$\vdots$}; }

\node[slot]  (a1) at (\cA, {0.90-\dy*0}) {\mdl{GPT OSS 120B}\ \ansW{17424}};
\node[slot]  (a2) at (\cA, {0.90-\dy*1}) {\mdl{GPT OSS 120B}\ \ansW{17424}};
\node[slot]  (a3) at (\cA, {0.90-\dy*2}) {\mdl{Gemma 3 12B IT}\ \ansW{132}};
\node[slot]  (b1) at (\cB, {0.90-\dy*0}) {\mdl{GPT OSS 120B}\ \ansW{263}};
\node[final] (c1) at (\cC, {0.90-\dy*0}) {\mdl{GPT OSS 120B}\ \ansR{69169}};

\draw[flow] (a1.east) -- (b1.west);
\draw[flow] (a2.east) -- (b1.west);
\draw[flow] (a3.east) -- (b1.west);
\draw[flow] (b1.east) -- (c1.west);
\draw[contrib] (pb1.east) -- (c1.west);
\draw[contrib] (pb2.east) -- (c1.west);

\foreach \s/\t in {a1/pb2, a3/pb1, pa3/pb1, pa3/pb3}
  { \draw[faint] (\s.east) -- (\t.west); }
\foreach \s/\t in {b1/pc2, pb2/pc1, pb3/pc1, pb3/pc3}
  { \draw[faint] (\s.east) -- (\t.west); }

\node[hdr] at (\cA, 1.15) {$S_1$};
\node[hdr] at (\cB, 1.15) {$S_2$};
\node[hdr] at (\cC, 1.15) {$S_3$};
\end{tikzpicture}}
\caption{Example of $M_\text{mix}$ escaping an attractor containing only incorrect answers by using a non-primary model to help revise its reasoning and reach the correct answer (69169).}
\label{app:fig:rsa}
\end{figure}

We present a hand-picked problem from IMO-AnswerBench that no individual model solved using either parallel or sequential scaling, but that recursive self-aggregation using $M_\text{mix}$ solved:

\begin{benchproblem}[title={IMO-AnswerBench}]
For a positive integer $n$, we call $g:\mathbb{Z}\rightarrow \mathbb{Z}$ a [sic] \textit{$n$-good function} if $g(1)=1$ and for any two distinct integers $a$ and $b$, $g(a)-g(b)$ divides $a^n -b^n$. We call a positive integer $n$ an \textit{exotic integer} if the number of $n$-good functions is twice of [sic] an odd integer. Find $132$th [sic] exotic integer.
\end{benchproblem}

Scaling methods using only GPT OSS 120B consistently return an incorrect answer of 17424. The derivation for this solution is correct except that it incorrectly includes some even candidates. The weaker Gemma 3 12B IT model often answers 132, which contains more errors, but approaches the problem from a different perspective. When combined in $M_\text{mix}$ with recursive self-aggregation, GPT OSS 120B aggregates these two solutions together (see \cref{app:fig:rsa}), forcing it to reconcile the two differing approaches. This leads GPT OSS 120B to identify the inconsistency with the answer of 17424 and generate a new answer, 263. This new solution is incorrect, but is corrected in the next state to obtain the correct answer of 69169. This example shows that, although the weaker Gemma model may not contribute much useful work itself, it can be used to pull a stronger model out of its local optimum.

\section{Breakdown of Problems Solved by Different Scaling Methods When Using $M_\text{mix}$ Compared to the Baseline}\label{sec:app:conf}

\begin{figure}
    \centering
    \includegraphics[width=0.75\linewidth]{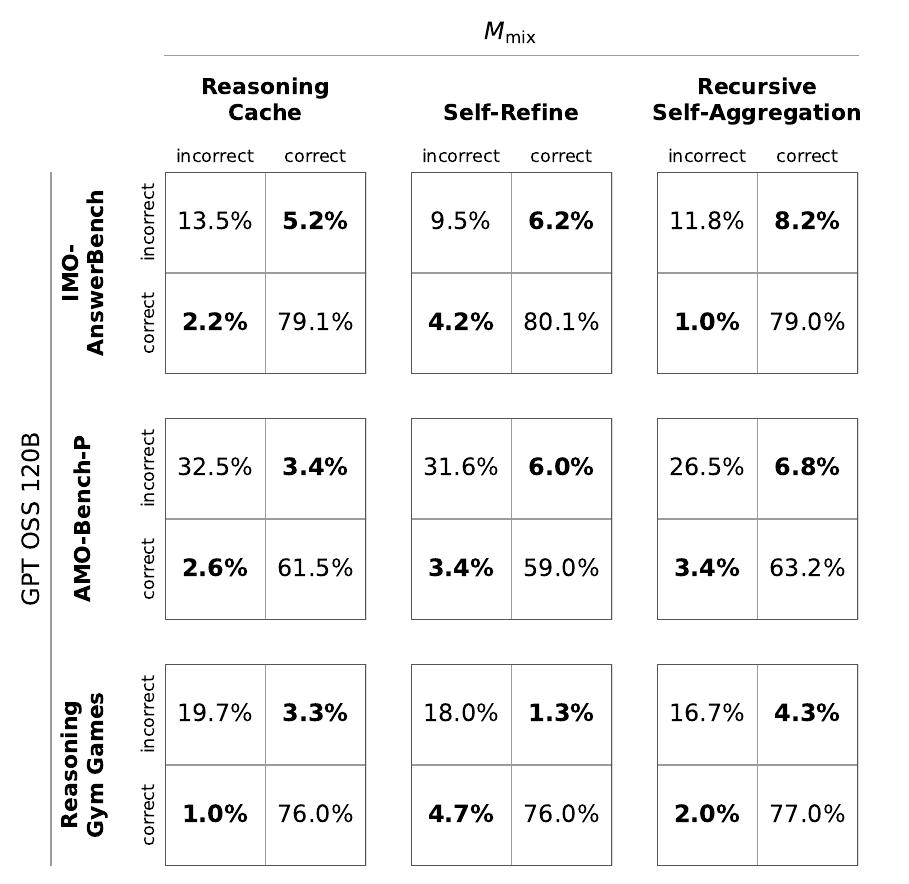}
    \caption{Per-problem coverage changes between $M_{\mathrm{mix}}$ and GPT OSS 120B with different sequential scaling methods. $M_\text{mix}$ improves over GPT OSS 120B without losing many correct answers.}
    \label{app:fig:conf}
\end{figure}

\Cref{mix:fig:exploration} shows that sequential scaling methods using $M_\text{mix}$ solve more problems than those using any single model. This implies there must be problems that only sequential scaling with $M_\text{mix}$ solves. \Cref{app:fig:conf} shows how many such problems there are. The problems solved by sequential scaling using $M_\text{mix}$ are relatively close to being supersets of the problems solved by the same methods with the primary model. This agrees with our hypothesis that correct answers are universal attractors across models, and therefore not lost when additional models are added to the mix.

\section{Cost of Using $M_\text{mix}$}\label{sec:app:cost}

\begin{figure}
    \centering
    \includegraphics[width=0.95\linewidth]{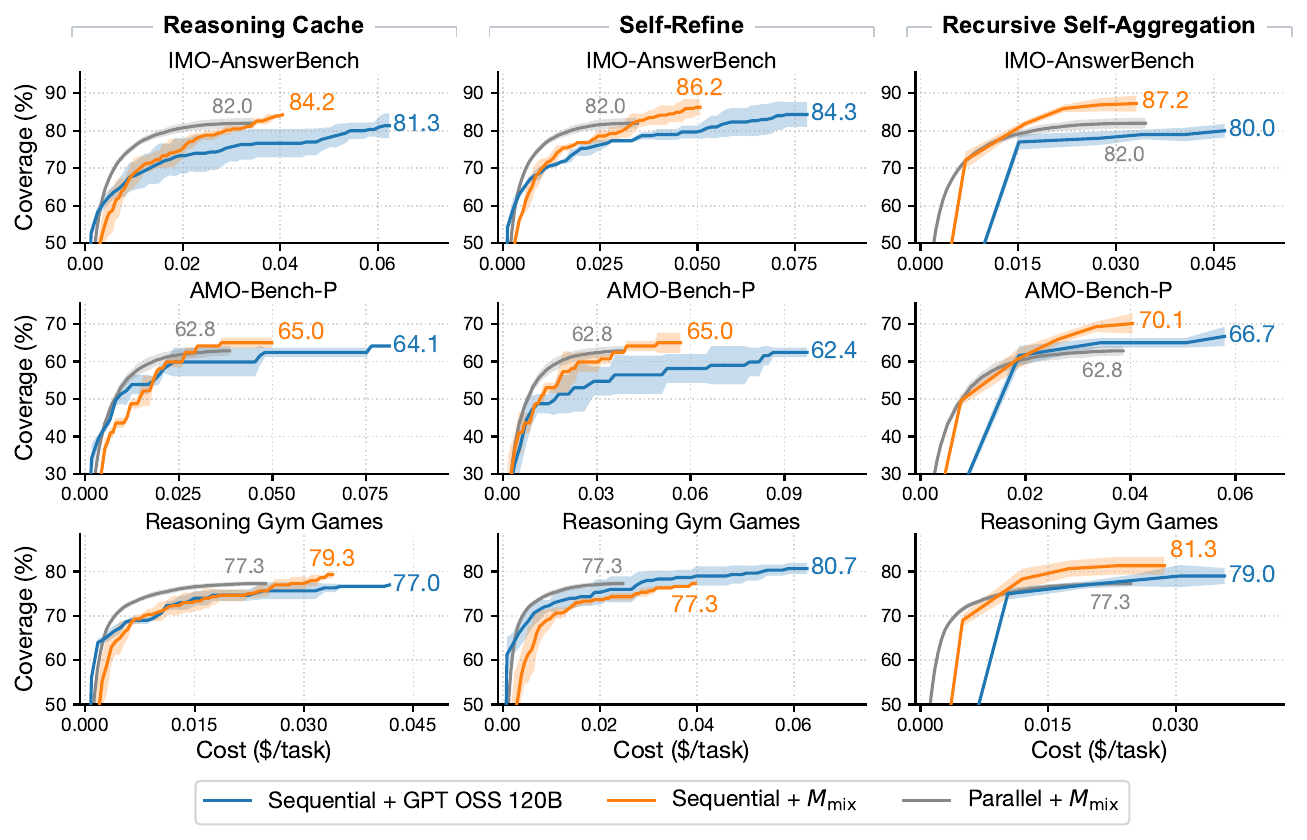}
    \caption{Coverage achieved by each scaling-method--model combination against price on \citet{openrouter_gpt_oss_120b} (identical to \cref{mix:fig:exploration}, but with cost on the x axis).}
    \label{app:fig:cost}
\end{figure}

\Cref{mix:fig:exploration} measures how coverage improves against the number of states visited because our focus is on the behaviour of the dynamical system. As a more practical baseline, \cref{app:fig:cost} presents the same coverage results against compute. To fairly compare compute, we report the cost of each trajectory using prices from \citet{openrouter_gpt_oss_120b}, assuming no cache hits. Although prices from \citet{openrouter_gpt_oss_120b} may be misleading, for example due to some models being subsidised, for the models we use, the prices also track with each model's reasoning capabilities.

\section{Additional Parallel Scaling Baselines}\label{sec:app:tot}

\begin{table}[t]
\centering
\caption{Parallel scaling coverage over 60 answers. The $M_\text{mix}$ row is identical to results displayed in \cref{mix:fig:exploration}. Sampling from $M_\text{mix}$ alone does not improve coverage over GPT OSS 120B, so its sequential scaling gains in \cref{mix:fig:exploration} are not explained by complementary model capabilities.}
\label{app:tab:parallel}
\vspace{2pt}
\begin{tabularx}{0.45\textwidth}{lYYY}
\toprule
& \multicolumn{3}{c}{Coverage (\%) $\uparrow$} \\
\cmidrule(lr){2-4}
Model & IMO & AMO & RGG \\
\midrule

GPT OSS 120B
& 82.8 & 64.5 & 79.0 \\

$M_\text{mix}$ (ours)
& 82.0 & 62.8 & 77.3 \\

\bottomrule
\end{tabularx}

\end{table}

\paragraph{$M_\text{mix}$ gains are not explained by complementary model capabilities.} In \cref{sec:mix:res}, we compare against a parallel baseline using $M_\text{mix}$. We do not use parallel scaling with only GPT OSS 120B in \cref{mix:fig:exploration} to ensure that improvements arise from additional exploration rather than from solutions contributed independently by the non-primary models, and because this comparison is unfair due to the baseline generating more answers with the primary model. Nonetheless, for transparency, we present this baseline in \cref{app:tab:parallel}. On all three benchmarks, GPT OSS 120B achieves higher coverage than $M_\text{mix}$, showing that adding answers from Gemma 3 12B IT and Granite 4.1 8B does not improve coverage enough to make up for the calls to GPT OSS 120B they replace. This justifies our selection of Gemma 3 12B IT and Granite 4.1 8B as models that offer very few capabilities on their own that are not already present in the primary model (in this case, GPT OSS 120B). Furthermore, sequential scaling with $M_\text{mix}$ still mostly achieves higher coverage than parallel scaling with only GPT OSS 120B.

\paragraph{The parallel baseline is not limited by redundant sampling.} A second possibility is that I.I.D. parallel sampling explores the unscaled models' answer distributions inefficiently, rather than the distribution itself being too restricted. We test this using tree-of-thoughts \citep{yao2023tree}, which prioritises generating \textit{diverse} and promising answers. We implement tree-of-thoughts on the Reasoning Gym Games benchmark with GPT OSS 120B (medium). At each round, we generate new answers using the `propose' recipe, with a branch factor of 3 in BFS mode. We evaluate states using the `value' recipe, averaged across two samples. At the end of four rounds, we return the resulting 60 answers. Across the 60 answers, the tree-of-thoughts implementation found at least one solution to 78.0\% of problems across three seeds, which is only slightly above the 77.3\% baseline coverage of I.I.D. parallel sampling in \cref{mix:fig:exploration}. We also investigate increasing temperature as a method for improving diversity in Appendix \ref{sec:app:temperature}, although this is already known to have a relatively small effect on reasoning performance \citep{renze2024effect}.

\section{Mixing Models Outperforms Other Perturbation Methods}\label{sec:app:temperature}

\begin{table}[t]
\centering
\caption{Performance of different sequential scaling methods with GPT-OSS-120B, when using the prompt-rephrasing method from \citet{wang2025effect}. This method does not significantly outperform the baseline.}
\label{app:tab:perturb}
\vspace{2pt}
\begin{tabularx}{0.75\textwidth}{llYYY}
\toprule

& &
\multicolumn{3}{c}{Coverage (\%) $\uparrow$} \\
\cmidrule(lr){3-5}
Method & Perturbation
& IMO & AMO & RGG \\
\midrule

\multirow{2}{*}{\parbox{2.8cm}{Reasoning Cache}}
& None & 81.3 & 64.1 & 77.0 \\
& \cite{wang2025effect} & 81.4 & 65.9 & 77.0 \\

\midrule

\multirow{2}{*}{\parbox{2.8cm}{Self-Refine}}
& None & 84.3 & 62.4 & 80.7 \\
& \cite{wang2025effect} & 83.1 & 62.6 & 79.3 \\

\midrule

\multirow{2}{*}{\parbox{2.8cm}{Recursive\\Self-Aggregation}}
& None & 80.0 & 66.7 & 79.0 \\
& \cite{wang2025effect} & 78.4 & 62.5 & 79.9 \\

\bottomrule
\end{tabularx}

\end{table}

Our proposed intervention perturbs the primary model to escape attractors by randomly switching to a secondary model. We propose this approach because attractors may appear due to biases in the model. In this section, we provide evidence that alternative approaches to introducing noise that do not change the underlying model -- such as perturbing the prompt or the sampling procedure -- do not produce the same improvements as our method.

\paragraph{Prompt-based noise.} One approach to introducing noise is to rephrase the prompt on each generation. We generate 100 rephrased descriptions of the problem using the primary model (\ie GPT-OSS-120B), with the same template as \citet{wang2025effect}. Before each response is generated, we substitute the original problem description in the prompt with a random rephrased version. \Cref{app:tab:perturb} shows the coverage achieved by prompt rephrasing with different sequential scaling methods, and suggests that rephrasing does not clearly improve performance over the baseline.

\begin{figure}
    \centering
    \includegraphics[width=0.45\linewidth]{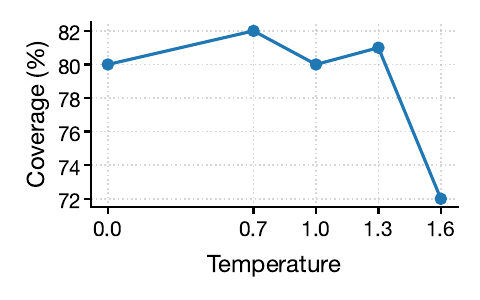}
    \caption{Coverage achieved by recursive self-aggregation with GPT OSS 120B (medium reasoning effort) on IMO-AnswerBench across different temperature parameters, using an identical configuration to \cref{mix:fig:exploration}.}
    \label{app:fig:temperature}
\end{figure}

\paragraph{Sampling-based noise} Another approach to introduce noise is to increase the temperature that models are sampled at. \Cref{app:fig:temperature} shows the coverage achieved by recursive self-aggregation on IMO-AnswerBench across different temperature parameters, using an identical configuration to \cref{mix:fig:exploration}. Changing the temperature from the value of 1 that we use elsewhere does not meaningfully improve coverage in these experiments. This result is similar to those of \citet{renze2024effect}.

\section{Omitted Ablation Results}\label{sec:app:fail}

\begin{figure}
    \centering
    \includegraphics[width=0.6\linewidth]{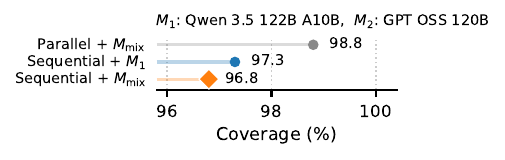}
    \caption{Results of mixing Qwen 3.5 122B A10B with GPT OSS 120B (both medium reasoning effort) in recursive self-aggregation, for IMO-AnswerBench. Except for the models used, this experiment is identical to those in \cref{mix:fig:ablations}.}
    \label{app:fig:fail}
\end{figure}

We do not include results in the main paper from experiments for which the baselines already saturated the dataset and therefore could not be meaningfully improved. For transparency, we present the experiment where this occurred in \cref{app:fig:fail}. We reject the experiment because the baselines score 395/400 and 389/400 so there is not sufficient space for our method to outperform the baseline: even a perfect score of 400/400 would not be a statistically significant improvement. We observe reduced performance because the better exploration our method enables cannot be demonstrated, while the disadvantage of our method using the primary model less often remains. To help demonstrate that the lack of improvement from our model-mixing intervention is due to the dataset being saturated for the baselines (and not because our results do not translate to larger models), \cref{intro:fig:main} shows that our model-mixing intervention is effective for other configurations of large models.

\section{Balancing Exploration and Exploitation with Annealed Model Sampling}\label{sec:app:ams}

\begin{figure}[t]
    \centering
    \includegraphics[width=0.95\linewidth]{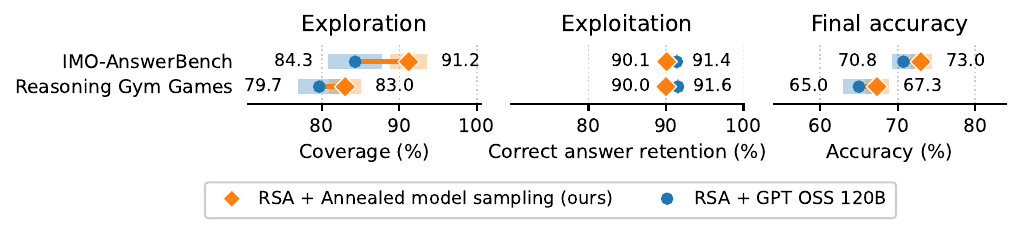}
    \caption{Comparison between recursive self-aggregation with a single model, and with annealed model sampling. For annealed model sampling, we initialise $c$ as 0.75 to select the model for the first answer, and then increase $c$ linearly at each attempt, until we have $c=1$ for the final attempt. We use an otherwise identical experimental setup to that in \cref{mix:fig:number} and display $\pm1$ standard deviation as the shaded region.}
    \label{mix:fig:final}
\end{figure}

While mixing models successfully improves exploration, it may also cause sequential scaling to escape the attractors of \textit{correct} answers. This could be true even given our hypothesis that correct answers are likely to remain as attractors for the mixture more often than incorrect ones will. In contrast, \cref{mix:fig:number} shows that using only the single best model leads to good exploitation (\ie the ability to retain good answers once we have found them), which aligns with prior results from \cite{li2024rethinking} that show that mixture-of-agents \citep{wang2024} often has reduced performance compared to using only the strongest model. However, there exist effective methods to transform high coverage into high accuracy \citep[e.g.][]{snell2024scaling,venkatraman2025recursive}. Therefore, to help demonstrate that escaping attractors can lead to improved overall performance, we present one such method that uses a simple heuristic to balance the increased exploration from model mixing with the stronger correct-answer exploitation from using just the primary model.

Let us assume $M_\text{mix}$ selects from three models, $\{M_1, M_2, M_3\}$, with $M_1$ being the primary model. We have suggested that sampling $j\in\{1,2,3\}$ with probability $[\frac{1}{3},\frac{1}{3},\frac{1}{3}]$ respectively (\ie a uniform distribution) leads to good exploration, and that sampling with probability $[1,0,0]$ (\ie using only the primary model) leads to good exploitation of correct answers that we have already found. Therefore we parametrise $M_\text{mix}$ by $c\in[0,1]$ (as in \cref{sec:mix:res}) to control the concentration of the sampling distribution $\frac{1}{3}\times[1+2c,1-c,1-c]$ ($0\le c\le1$) over time. We propose to use a distribution with good exploration to generate our initial states, and then move towards better exploitation for later states. This is similar to simulated annealing, where initial states search for a region of the answer space that contains a good attractor, while later states more carefully converge to its centre.

\Cref{mix:fig:final} compares recursive self-aggregation \citep{venkatraman2025recursive} (\ie with $c=1$) to the same method, but using $c=0.75$ (the best exploration setting in \cref{mix:fig:number}) to select the model for the first answer, $c=1$ (the best exploitation setting in \cref{mix:fig:number}) for the final answer, and a linear interpolation for the answers in between. This annealed model sampling achieves higher accuracy than the baseline recursive self-aggregation in both benchmarks by improving its exploration while achieving similar exploitation performance. Furthermore, because annealed model sampling sometimes swaps the primary model for weaker -- and cheaper -- models, its cost on \citet{openrouter_gpt_oss_120b} is 3.6\% and 3.4\% lower than that of the baseline for IMO-AnswerBench and Reasoning Gym Games respectively.

\section{Results on ARC AGI 1}\label{sec:app:arc}

\begin{table}
\centering
\caption{Attractor dynamics on ARC AGI 1, similar to \cref{attractors:tab:part_1,mix:tab:part_2}. Sequential scaling is less likely to get stuck in attractors within the first few states compared to the other benchmarks tested.}
\label{app:tab:arc}
\vspace{2pt}
\begin{tabularx}{\textwidth}{llYY}
\toprule
Method & Model
& Attractor hit by $S_4$ (\%) $\downarrow$
& Different attractors (\%) $\downarrow$ \\
\midrule

\multirow{4}{*}{\parbox{2.8cm}{Reasoning Cache}}
& GPT OSS 120B
& 20.6 & 23.6 \\
& Gemma 3 12B IT
& \textbf{13.3} & \textbf{16.4} \\
& Granite 4.1 8B
& 25.9 & 53.4 \\
& \hl$M_\text{mix}$ (ours)
& \hl16.4 & \hl34.8 \\

\midrule

\multirow{4}{*}{\parbox{2.8cm}{Self-Refine}}
& GPT OSS 120B
& 8.0 & \textbf{25.0} \\
& Gemma 3 12B IT
& \textbf{7.7} & 30.0 \\
& Granite 4.1 8B
& 15.0 & 30.0 \\
& \hl$M_\text{mix}$ (ours)
& \hl12.7 & \hl26.7 \\

\midrule

\multirow{4}{*}{\parbox{2.8cm}{Recursive\\Self-Aggregation}}
& GPT OSS 120B
& 26.3 & 5.3 \\
& Gemma 3 12B IT
& 22.6 & \textbf{3.0} \\
& Granite 4.1 8B
& \textbf{22.4} & 5.9 \\
& \hl$M_\text{mix}$ (ours)
& \hl24.3 & \hl4.1 \\

\bottomrule
\end{tabularx}

\end{table}

\begin{figure}[t!]
    \centering
    \includegraphics[width=0.95\linewidth]{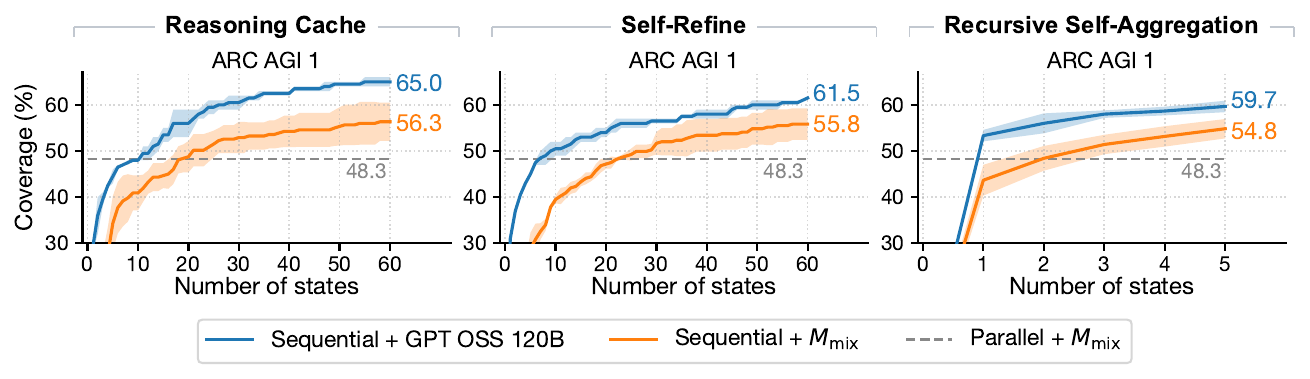}
    \caption{Coverage over time on ARC AGI 1 (similar to \cref{mix:fig:exploration}). The GPT OSS 120B baseline continues improving after the first few states, outperforming $M_\text{mix}$.}
    \label{app:fig:arc_cover}
\end{figure}

\begin{figure}
    \centering
    \includegraphics[width=0.75\linewidth]{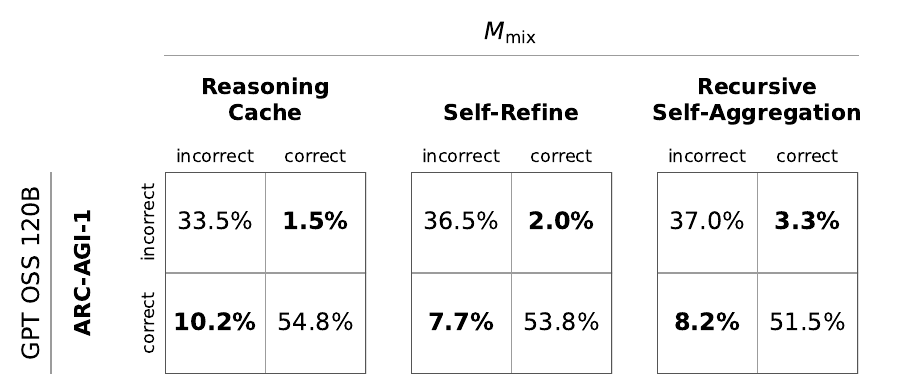}
    \caption{Per-problem agreement between $M_{\mathrm{mix}}$ and GPT OSS 120B on ARC AGI 1 (similar to \cref{app:fig:conf}).}
    \label{app:fig:arc_num}
\end{figure}

We repeat our experiments from sections \ref{sec:attractors} and \ref{sec:mix} for the ARC AGI 1 dataset \citep{chollet2019measure}. \Cref{app:tab:arc} shows that, before any intervention with $M_\text{mix}$, the sequential scaling methods tested get stuck in attractors less often than for other datasets, and, when they are stuck, the attractor they find is more often the same across multiple different trajectories. This implies there are likely fewer attractors, and larger basins, rather than many small, suboptimal attractors. As a result, the baseline using only GPT OSS 120B outperforms our proposed method using $M_\text{mix}$ (Figures \ref{app:fig:arc_cover} and \ref{app:fig:arc_num}). This is because the baseline continues to improve for longer compared to the other datasets, so mixing models offers little benefit while reducing how often the strongest model is used.

Why does ARC AGI 1 not hit attractors in the first four states as often as the other datasets? A difference between the datasets in Sections \ref{sec:attractors} and \ref{sec:mix}, and ARC AGI 1 is that the latter does not have an objective that the model can attempt to verify. Specifically, problems based on maths and games provide a clear goal that the language model can use to test whether it has succeeded. This can lead to suboptimal attractors, because the language model may be unable to spot a logical error in its reasoning, and therefore conclude that the answer it has already obtained must be correct. This is not only true for verifiable problems: a task where we are asked to summarise a piece of text, for example, may define goals of faithfulness to the text and brevity of the summary. A language model could receive a summary and conclude that it has already maximised both of these goals. Conversely, ARC AGI 1 displays patterns and then asks the model to decide what comes next, but defines no clear rules that the language model can use to argue that a given answer must be correct, potentially making it harder for sequential scaling to get stuck in suboptimal attractors. A more in-depth exploration of exactly which tasks attractors appear in would be costly, so we leave this for future work.

\end{document}